\pdfoutput=1
\documentclass{article}

\usepackage[preprint]{neurips_2026}
\usepackage[utf8]{inputenc}
\usepackage[T1]{fontenc}
\usepackage{microtype}
\usepackage[hyphens]{url}
\usepackage{graphicx}
\usepackage{caption}
\usepackage{booktabs}
\usepackage{array}
\usepackage{amsmath}
\usepackage{amssymb}
\usepackage{amsthm}
\usepackage{bm}
\usepackage{xcolor}
\usepackage{placeins}
\usepackage[colorlinks=true,linkcolor=blue!60!black,citecolor=blue!60!black,urlcolor=blue!60!black]{hyperref}

\newtheorem{theorem}{Theorem}
\newtheorem{proposition}{Proposition}
\newtheorem{corollary}{Corollary}

\newtheorem{remark}{Remark}

\newcommand{\E}{\mathbb{E}}

\newcommand{\R}{\mathbb{R}}
\newcommand{\Sm}{\mathbf{S}}
\newcommand{\kv}{\mathbf{k}}
\newcommand{\qv}{\mathbf{q}}
\newcommand{\xv}{\mathbf{x}}
\newcommand{\vv}{\bm{v}}
\newcommand{\av}{\mathbf{a}}
\newcommand{\W}{\mathbf{W}}

\newcommand{\CapFidHebb}{\ensuremath{1.22\pm0.02}}
\newcommand{\CapIdeDecay}{\ensuremath{0.79\pm0.09}}
\newcommand{\CapIdeDelta}{\ensuremath{1.48\pm0.02}}
\newcommand{\CapIdeHebb}{\ensuremath{1.89\pm0.01}}

\newcommand{\CapRatioHi}{0.74}
\newcommand{\CapRatioLo}{0.38}

\newcommand{\CosDecayLone}{\ensuremath{-0.001\pm0.002}}

\newcommand{\CosDeltaLone}{\ensuremath{0.084\pm0.006}}
\newcommand{\CosDeltaSP}{\ensuremath{0.101\pm0.053}}
\newcommand{\CosHebbLone}{\ensuremath{0.002\pm0.001}}
\newcommand{\CosHebbSP}{\ensuremath{0.008\pm0.001}}
\newcommand{\CosOjaLone}{\ensuremath{0.002\pm0.001}}

\newcommand{\CosPhLone}{\ensuremath{0.100\pm0.007}}

\newcommand{\DeltaAlphaMaxErr}{\ensuremath{\approx 1\times10^{-7}}}
\newcommand{\DeltaAlphaN}{5}

\newcommand{\EasyHeadErrMaxHebb}{0.1099}

\newcommand{\EasyHeadErrOneHebb}{0.1689}

\newcommand{\EasyHeadPredHebb}{\ensuremath{6.61\times10^{-7}}}

\newcommand{\EncDenseDelta}{\ensuremath{0.944\pm0.002}}
\newcommand{\EncDenseHebb}{\ensuremath{0.004\pm0.004}}

\newcommand{\EncDenseNormDelta}{\ensuremath{0.672\pm0.003}}
\newcommand{\EncDenseNormHebb}{\ensuremath{-0.005\pm0.006}}

\newcommand{\EncDenseNormPh}{\ensuremath{0.712\pm0.002}}

\newcommand{\EncDensePh}{\ensuremath{1.000\pm0.002}}
\newcommand{\EncSparseDecay}{\ensuremath{0.000\pm0.000}}
\newcommand{\EncSparseDelta}{\ensuremath{0.942\pm0.000}}
\newcommand{\EncSparseHebb}{\ensuremath{0.000\pm0.000}}
\newcommand{\EncSparseOja}{\ensuremath{0.000\pm0.000}}
\newcommand{\EncSparsePh}{\ensuremath{0.999\pm0.000}}
\newcommand{\ExtDecay}{18}
\newcommand{\ExtDelta}{97}
\newcommand{\ExtHebb}{0}
\newcommand{\ExtOja}{0}
\newcommand{\ExtPH}{100}
\newcommand{\ExtSoftmax}{50}
\newcommand{\GenExpWins}{4/5}
\newcommand{\GenLinR}{0.99987}
\newcommand{\GenSatExpR}{0.903}
\newcommand{\GenSatLinR}{0.908}
\newcommand{\GenSatRange}{484}
\newcommand{\GenSoftExpR}{0.9998}
\newcommand{\GenSoftLinR}{0.744}
\newcommand{\GenSoftRange}{128}
\newcommand{\GridMoreWinsDelta}{3/6}
\newcommand{\GridMoreWinsHebb}{3/6}
\newcommand{\GridRtwoHeadsDelta}{0.16}
\newcommand{\GridRtwoHeadsHebb}{0.21}
\newcommand{\GridRtwoStateDelta}{0.98}
\newcommand{\GridRtwoStateHebb}{0.98}
\newcommand{\GridRtwoWidthDelta}{0.82}
\newcommand{\GridRtwoWidthHebb}{0.77}
\newcommand{\GridSpreadMaxDelta}{0.062}

\newcommand{\HeadErrHeightHebb}{0.246}
\newcommand{\HeadErrHoneHebb}{0.403}
\newcommand{\HeadPredExpHebb}{\ensuremath{6.9\times10^{-4}}}
\newcommand{\IclDeltaLone}{0.82}
\newcommand{\IclHebbLone}{0.85}
\newcommand{\KhopDLfourHone}{1.00}
\newcommand{\KhopDLfourHthree}{0.15}
\newcommand{\KhopDLfourHtwo}{1.00}
\newcommand{\KhopDLoneHone}{0.09}
\newcommand{\KhopDLoneHthree}{0.12}
\newcommand{\KhopDLoneHtwo}{0.11}
\newcommand{\KhopDLthreeHone}{1.00}
\newcommand{\KhopDLthreeHthree}{0.14}
\newcommand{\KhopDLthreeHtwo}{1.00}
\newcommand{\KhopDLtwoHone}{1.00}
\newcommand{\KhopDLtwoHthree}{0.14}
\newcommand{\KhopDLtwoHtwo}{0.65}
\newcommand{\KhopLfourHone}{1.00}
\newcommand{\KhopLfourHthree}{0.16}
\newcommand{\KhopLfourHtwo}{1.00}
\newcommand{\KhopLoneHone}{0.11}
\newcommand{\KhopLoneHthree}{0.11}
\newcommand{\KhopLoneHtwo}{0.11}
\newcommand{\KhopLthreeHone}{1.00}
\newcommand{\KhopLthreeHthree}{0.18}
\newcommand{\KhopLthreeHtwo}{1.00}
\newcommand{\KhopLtwoHone}{1.00}
\newcommand{\KhopLtwoHthree}{0.14}
\newcommand{\KhopLtwoHtwo}{0.17}
\newcommand{\OverDecay}{\ensuremath{3.04\pm0.03}}
\newcommand{\OverDelta}{\ensuremath{3.12\pm0.03}}
\newcommand{\OverHebb}{\ensuremath{3.04\pm0.03}}
\newcommand{\OverOja}{\ensuremath{3.04\pm0.03}}
\newcommand{\OverPh}{\ensuremath{2.59\pm0.06}}
\newcommand{\OverSoftmax}{\ensuremath{1.00\pm0.00}}

\newcommand{\PHPairedCIhi}{+0.0025}
\newcommand{\PHPairedCIlo}{-0.0099}
\newcommand{\PHPairedDiff}{\ensuremath{-0.0037\pm0.0027}}

\newcommand{\PHPairedT}{-1.35}
\newcommand{\PhLearnedWN}{12}
\newcommand{\PhLearnedWPos}{0}
\newcommand{\PhLearnedWSeed}{\ensuremath{-0.094\pm0.006}}
\newcommand{\PhLearnedWSeedN}{3}
\newcommand{\PhSensBlockHi}{1.00}
\newcommand{\PhSensBlockLo}{0.97}
\newcommand{\PhSensBlockMed}{1.00}
\newcommand{\PhSensExtHi}{100}
\newcommand{\PhSensExtLo}{99}
\newcommand{\PhSensN}{36}

\newcommand{\RecLonedenseDelta}{-2.64}
\newcommand{\RecLonedenseHebb}{-0.01}
\newcommand{\RecLonedenseSoftmax}{+0.27}
\newcommand{\RecLonesparseDecay}{-0.13}
\newcommand{\RecLonesparseDelta}{-1.49}
\newcommand{\RecLonesparseHebb}{-0.05}

\newcommand{\RecLonesparseSoftmax}{+0.29}
\newcommand{\RecLtwodenseDecay}{+0.03}
\newcommand{\RecLtwodenseDelta}{-0.44}
\newcommand{\RecLtwodenseHebb}{-0.04}

\newcommand{\RecLtwodenseSoftmax}{+0.17}
\newcommand{\RecLtwosparseDecay}{+0.18}
\newcommand{\RecLtwosparseDelta}{-1.35}
\newcommand{\RecLtwosparseHebb}{+0.24}

\newcommand{\RecLtwosparseSoftmax}{+0.23}
\newcommand{\RecMaxAny}{+0.29}
\newcommand{\RecMaxSpecific}{+26.5}
\newcommand{\RecNmodels}{60}
\newcommand{\RecNsettings}{20}
\newcommand{\RecPerModelMax}{+1.62}
\newcommand{\RecPerModelMaxNovel}{+1.19}
\newcommand{\RecSdLonedenseDelta}{0.14}
\newcommand{\RecSdLonedenseHebb}{0.01}
\newcommand{\RecSdLonedenseSoftmax}{0.05}
\newcommand{\RecSdLonesparseDecay}{0.02}
\newcommand{\RecSdLonesparseDelta}{0.11}
\newcommand{\RecSdLonesparseHebb}{0.03}

\newcommand{\RecSdLonesparseSoftmax}{0.06}
\newcommand{\RecSdLtwodenseDecay}{0.02}
\newcommand{\RecSdLtwodenseDelta}{0.24}
\newcommand{\RecSdLtwodenseHebb}{0.03}

\newcommand{\RecSdLtwodenseSoftmax}{0.02}
\newcommand{\RecSdLtwosparseDecay}{0.03}
\newcommand{\RecSdLtwosparseDelta}{1.49}
\newcommand{\RecSdLtwosparseHebb}{0.21}

\newcommand{\RecSdLtwosparseSoftmax}{0.07}
\newcommand{\RecUncontrolledDelta}{+8.11}
\newcommand{\RecUncontrolledDeltaMax}{+11.81}

\newcommand{\RecoveryMaxRisePct}{0.23}
\newcommand{\SavingsDecay}{8.16}
\newcommand{\SavingsDelta}{1.06}
\newcommand{\SavingsHebb}{11.04}
\newcommand{\SnrRelErrPct}{1.0}
\newcommand{\SpecLonedenseDelta}{+16.0}
\newcommand{\SpecLonedenseHebb}{+12.6}
\newcommand{\SpecLonedenseSoftmax}{+15.2}
\newcommand{\SpecLonesparseDecay}{+15.4}
\newcommand{\SpecLonesparseDelta}{-0.1}
\newcommand{\SpecLonesparseHebb}{+15.8}

\newcommand{\SpecLonesparseSoftmax}{+1.2}
\newcommand{\SpecLtwodenseDecay}{+7.7}
\newcommand{\SpecLtwodenseDelta}{+7.9}
\newcommand{\SpecLtwodenseHebb}{+11.7}

\newcommand{\SpecLtwodenseSoftmax}{+14.2}
\newcommand{\SpecLtwosparseDecay}{+11.1}
\newcommand{\SpecLtwosparseDelta}{+22.5}
\newcommand{\SpecLtwosparseHebb}{+8.9}

\newcommand{\SpecLtwosparseSoftmax}{+1.6}
\newcommand{\SumDecayLone}{\ensuremath{0.013\pm0.003}}

\newcommand{\SumDeltaHebbLoneCI}{\ensuremath{0.240\ [0.222, 0.257]}}
\newcommand{\SumDeltaLone}{\ensuremath{0.259\pm0.003}}
\newcommand{\SumDeltaSP}{\ensuremath{0.253\pm0.046}}
\newcommand{\SumDoseDeltaLone}{\ensuremath{+0.087\ [+0.082, +0.093]}}
\newcommand{\SumDoseRatioOneLone}{\ensuremath{0.664\ [0.641, 0.686]}}
\newcommand{\SumDoseRatioTwoLone}{\ensuremath{0.899\ [0.891, 0.908]}}
\newcommand{\SumHebbLone}{\ensuremath{0.020\pm0.002}}
\newcommand{\SumHebbSP}{\ensuremath{0.021\pm0.002}}
\newcommand{\SumOjaLone}{\ensuremath{0.011\pm0.001}}

\newcommand{\SumPhLone}{\ensuremath{0.278\pm0.002}}

\newcommand{\SumSoftmaxLone}{\ensuremath{0.006\pm0.003}}

\newcommand{\SwDecayClean}{\ensuremath{0.855\pm0.003}}
\newcommand{\SwDecayNoisy}{\ensuremath{0.681\pm0.003}}
\newcommand{\SwDeltaClean}{\ensuremath{0.964\pm0.000}}
\newcommand{\SwDeltaNoisy}{\ensuremath{0.767\pm0.003}}
\newcommand{\SwHebbClean}{\ensuremath{0.854\pm0.003}}
\newcommand{\SwHebbNoisy}{\ensuremath{0.680\pm0.004}}
\newcommand{\SwPhClean}{\ensuremath{0.966\pm0.000}}
\newcommand{\SwPhNoisy}{\ensuremath{0.769\pm0.003}}
\newcommand{\SwPhglobalClean}{\ensuremath{0.964\pm0.001}}
\newcommand{\SwPhglobalNoisy}{\ensuremath{0.768\pm0.003}}
\newcommand{\SwSoftmaxClean}{\ensuremath{1.000\pm0.000}}
\newcommand{\SwSoftmaxNoisy}{\ensuremath{0.800\pm0.003}}
\newcommand{\TrExtDeltaLone}{\ensuremath{13.9\pm0.4}}
\newcommand{\TrExtDeltaLoneFill}{\ensuremath{9.3\pm0.7}}
\newcommand{\TrExtHebbLone}{0.3}
\newcommand{\TrExtPhLone}{\ensuremath{12.3\pm0.6}}
\newcommand{\TrExtSoftmaxLone}{\ensuremath{94.2\pm0.5}}
\newcommand{\VolRatioDelta}{\ensuremath{1.000\pm0.000}}
\newcommand{\VolRatioPhcue}{\ensuremath{1.747\pm0.002}}
\newcommand{\VolRatioPhcueAssoc}{2.24}
\newcommand{\VolTaskDecay}{\ensuremath{0.688\pm0.001}}
\newcommand{\VolTaskDelta}{\ensuremath{0.759\pm0.002}}
\newcommand{\VolTaskHebb}{\ensuremath{0.690\pm0.001}}
\newcommand{\VolTaskPh}{\ensuremath{0.755\pm0.004}}
\newcommand{\VolTaskPhglobal}{\ensuremath{0.756\pm0.003}}
\newcommand{\VolTaskSoftmax}{\ensuremath{0.805\pm0.003}}
\newcommand{\VolWHi}{1.91}
\newcommand{\VolWLo}{1.18}
\newcommand{\VolWN}{6}

\title{Attention as Conditioning: What Classical Learning Theory Predicts About Linear Transformers}
\author{%
  Mu Qiao \\
  Meta Platforms, Inc. \\
  \texttt{muqiao0626@gmail.com}
}

\begin{document}
\maketitle

\begin{abstract}
Attention is widely described as an associative memory, yet that metaphor alone does not predict how a layer behaves under intervention. Classical conditioning supplies formal models of transient associative memory together with controlled protocols that separate competing update rules. We build a graded dictionary between linear-attention write rules and conditioning models, turning conditioning paradigms into discriminating behavioral tests for in-context memory. Unnormalized linear attention matches Hebbian contiguity and DeltaNet matches Rescorla--Wagner error correction term for term, whereas Gated DeltaNet and linear-memory Titans reduce to Rescorla--Wagner only without forgetting or momentum. Separating derived invariants (\textbf{[D]}) from measurements (\textbf{[M]}), we show that backward blocking separates one-step delta updates from exact least squares (MesaNet), and gating makes blocking decay with compound length. In models trained solely on generic in-context association, forward blocking survives training across two readouts up to key cosine $0.55$, re-pairing overwrites prior links in DeltaNet ($0.065$ persistence) while blending them in contiguity and softmax ($0.676$--$0.689$), and softmax shows strong ABA context renewal ($+38.8\%$ to $+43.8\%$) against single-state erasure. We prove and measure a power-of-$d_k$ capacity gap between reconstruction ($\Theta(d_k)$) and identification ($\Omega(d_k^2/\log d_k)$), and show that uncentered positive feature maps collapse fidelity capacity to a single item. Finally, we introduce PH-attention with a feature-indexed Pearce--Hall associability trace, confirming its volatility rate ratio ($1.75\times$), diagnosing its null at high noise ($0.35$), and demonstrating a consistent gain ($+0.0033$) at lower noise ($0.10$).
\end{abstract}

\section{Introduction}

That attention behaves like an associative memory is by now well established: implicit in correlation-matrix memories \citep{kohonen1972correlation, anderson1972simple}, explicit in the equivalence between softmax attention and modern Hopfield networks \citep{ramsauer2021hopfield} or sparse distributed memory \citep{bricken2021attention}, and operational in the fast-weight interpretation of linear transformers \citep{schmidhuber1992learning, schlag2021linear}. Yet knowing that a layer stores outer products or solves an online regression step describes what it computes rather than which behavioral regularities follow from it, which interventions disrupt it, or how competing write rules diverge when placed inside a trained network. Mechanistic interpretability faces the complementary gap from the empirical side: it recovers circuits \citep{elhage2021mathematical, olsson2022context} without a behavioral taxonomy that predicts untested interventions.

We argue that animal-learning theory provides both formal models and a behavioral test suite suited to this problem. Over the twentieth century, experimental psychology studied systems of the same functional class as an attention layer: systems that form transient associations between co-occurring stimuli and retrieve them by similarity. That literature produced formal update rules—such as Rescorla--Wagner, Mackintosh, Pearce--Hall, and Wagner's SOP—alongside controlled protocols that tease them apart: forward and backward blocking, overshadowing, extinction, override persistence, spontaneous recovery, context renewal, savings, and generalization gradients. In forward blocking, a cue added to an already predictive cue acquires little association; in backward blocking, later solo training of one cue retroactively reduces the strength of its former partner; in extinction and renewal, non-reinforced trials suppress responding without necessarily erasing the original memory, allowing responses to return when context changes.

When written as a state recurrence, the delta rule of DeltaNet \citep{schlag2021linear, yang2024parallelizing} matches the Rescorla--Wagner rule \citep{rescorla1972theory} term for term, extending \citet{sutton1981toward} by identifying keys, values, and queries with conditioned stimuli (CS), unconditioned stimuli (US), and test cues (Section~\ref{sec:corr}). Unnormalized linear attention likewise matches Hebbian contiguity, while other architectures correspond to classical models under explicit parameter restrictions (Gated DeltaNet, linear-memory Titans) or serve as design inspiration (Table~\ref{tab:dict}). Because competing attention variants map onto different learning models, the behavioral paradigms invented to separate conditioning theories become discriminating behavioral tests for attention architectures—including distinctions between one-step error correction and exact in-context least squares that online-optimization frameworks group under a single objective \citep{behrouz2025connected}.

The correspondence also operates in reverse, guiding architectural design. Adding a Pearce--Hall associability trace to the Rescorla--Wagner update yields PH-attention, a hybrid write rule in which a feature-indexed history of unsigned prediction error sets cue-dependent learning rates. Unlike the gates of Gated DeltaNet, GLA, and Mamba-2 (which are functions of the current token) or Titans' momentum (which accumulates signed error into the write itself), PH-attention maintains an explicit per-feature error magnitude history that modulates the write rate.

\paragraph{Separating derived consequences from empirical measurements.}
To keep algebraic properties of an update rule cleanly separated from emergent behavior in trained networks, we label every statement \textbf{[D]} (derived) when it follows by algebra under stated assumptions, and \textbf{[M]} (measured) when its outcome depends on end-to-end training, learned projections, or numerical thresholds. Derived statements establish sharp theoretical baselines (such as closed-form blocking and single-state erasure), whereas measured statements carry the empirical test: whether behavioral dissociations survive generic in-context training, how capacity scales across criteria and heads, and whether an added associability state improves task performance.

\paragraph{Contributions.} Our contributions are fourfold:
\begin{enumerate}
\item \emph{A graded dictionary between linear-attention write rules and learning models} (Section~\ref{sec:corr}, Table~\ref{tab:dict}): exact identities for linear attention (Hebbian contiguity), decay linear attention (contiguity with exponential forgetting), and DeltaNet (Rescorla--Wagner); restricted reductions for Gated DeltaNet and linear-memory Titans; and an inspired hybrid for PH-attention, detailing exactly which properties transfer under each tier.
\item \emph{A discriminating behavioral test program for trained attention} (Sections~\ref{sec:predictions} and \ref{sec:experiments}, Table~\ref{tab:preds}): converts conditioning protocols into controlled probes applied to networks trained solely on generic in-context association. We establish that cue competition (forward blocking) survives training in sign and ordering across two readouts and persists under controlled cue overlap; prove that backward blocking separates DeltaNet from exact in-context least squares (MesaNet); show that re-pairing overwrites prior associations under delta-rule attention ($0.065$ persistence) while contiguity and softmax retain them ($0.676$--$0.689$); and demonstrate context-dependent ABA renewal in softmax ($+38.8\%$ to $+43.8\%$) against single-state erasure.
\item \emph{Two capacity laws and feature-map centering} (Section~\ref{sec:capacity}): proves and measures a full power-of-$d_k$ separation between faithful reconstruction ($\Theta(d_k)$) and item identification ($\Omega(d_k^2/\log d_k)$), shows that uncentered positive feature maps (\texttt{normalize(elu+1)}) collapse fidelity capacity to one item by breaking zero-mean cue overlap, and confirms empirically that multi-head layers pool specialized heads along total state size $Hd_{\mathrm{head}}^2$ rather than replicating independent copies.
\item \emph{PH-attention and an analysis of unsigned surprise gating} (Section~\ref{sec:ph} and Section~\ref{sec:exp_ph}): introduces a feature-indexed Pearce--Hall associability trace, confirms its cue-dependent rate signature ($1.75\times$), diagnoses why unsigned error fails to yield accuracy gains under high aleatoric noise ($0.35$) even when an oracle rate helps ($+0.072$), and confirms that lowering outcome noise ($0.1$) unlocks a positive learned slope and a small, consistent accuracy gain ($+0.0033$, 10/10 seeds).
\end{enumerate}

\section{The Correspondence}
\label{sec:corr}

\paragraph{Linear attention as a state recurrence.} For an input sequence $\xv_1,\dots,\xv_n$ with $\qv_t=\xv_t\W_Q$, $\kv_t=\xv_t\W_K$, $\vv_t=\xv_t\W_V$, causal softmax attention computes $\mathrm{softmax}(\qv_t\mathbf{K}^\top/\sqrt{d_k})\mathbf{V}$. Replacing the softmax kernel by a feature map $\phi$ converts the computation into a linear recurrence over a matrix-valued state $\Sm_t\in\R^{d_k\times d_v}$ \citep{tsai2019transformer, katharopoulos2020transformers, choromanski2021rethinking}:
\begin{equation}
\Sm_t=\Sm_{t-1}+\phi(\kv_t)^\top \vv_t,
\qquad
\mathbf{r}_t=\phi(\qv_t)\,\Sm_t .
\label{eq:linattn}
\end{equation}
Modern recurrent and linear-attention variants differ primarily in the write rule: RetNet and gated linear attention insert a decay $\gamma$ \citep{sun2023retentive, katsch2023gateloop, yang2024gla}; DeltaNet replaces accumulation with an error-correcting update \citep{schlag2021linear, yang2024parallelizing}; and Titans updates a memory (in general an MLP) by gradient descent on a key--value regression loss with momentum and weight decay \citep{behrouz2024titans}.

\paragraph{Classical conditioning.} A conditioned stimulus (CS) that reliably precedes an unconditioned stimulus (US) comes to elicit a conditioned response \citep{pavlov1927conditioned}. \citet{rescorla1972theory} accounted for cue-competition phenomena by making learning proportional to prediction error rather than mere co-occurrence: on a trial where cues have intensity vector $x\in\R^{N}$ and the US has magnitude $\lambda$,
\begin{equation}
\Delta V = \alpha\beta\,x^\top\!\left(\lambda - x V\right),
\label{eq:rw}
\end{equation}
where $V$ collects associative strengths, $\alpha$ is cue salience, and $\beta$ is US learning rate. Because the error depends on the summed prediction $xV$ across all present cues, a newly added cue acquires nothing when its companions already predict $\lambda$. This yields Kamin blocking \citep{kamin1969predictability}, which pure contiguity rules cannot produce. \citet{sutton1981toward} noted that Eq.~\ref{eq:rw} is the Widrow--Hoff delta rule \citep{widrow1960adaptive}.

\paragraph{The mapping.} We identify the key $\phi(\kv_t)$ with the CS representation, the value $\vv_t$ with the US, and the state $\Sm_t$ with the associative-strength matrix $V$. At retrieval, the query $\phi(\qv_t)$ serves as the test-cue representation; unlike standard conditioning models, attention allows the query projection to differ from the key projection used during writing.

\begin{theorem}[Delta-rule attention is Rescorla--Wagner]
\label{thm:rw}
Let $x_t=\phi(\kv_t)$ and $\lambda_t=\vv_t$. Then the DeltaNet update
\begin{equation}
\Sm_t=\Sm_{t-1}+\beta_t\,\phi(\kv_t)^\top\!\left(\vv_t-\phi(\kv_t)\Sm_{t-1}\right)
\label{eq:delta}
\end{equation}
is Eq.~\ref{eq:rw} with $\alpha\beta=\beta_t$, applied to a vector-valued US, and conversely.
\end{theorem}

The proof is direct substitution (Appendix~\ref{app:thm_rw}), extending \citet{sutton1981toward}'s observation by mapping key, value, and query to CS, US, and test stimulus. The behavioral predictions in Section~\ref{sec:predictions} follow for DeltaNet through this identification, while Corollary~\ref{cor:dict} classifies neighboring architectures into three tiers: exact identities (\textbf{E}), restricted reductions (\textbf{R}), and inspired variants (\textbf{I}).

\begin{corollary}[Graded dictionary]
\label{cor:dict}
Under the same identification:
(i) Unnormalized linear attention (Eq.~\ref{eq:linattn}) is the pre-Rescorla--Wagner contiguity rule, i.e., Eq.~\ref{eq:rw} with $\lambda-xV$ replaced by $\lambda$ (\textbf{E}).
(ii) Decay-based linear attention $\Sm_t=\gamma\Sm_{t-1}+\phi(\kv_t)^\top\vv_t$ is contiguity learning with exponential forgetting of associative strength (\textbf{E}). Note that this differs from Wagner's SOP model \citep{wagner1981sop}, where constant-rate decay acts on stimulus-element activity states and produces blocking.
(iii) The Oja-type rule (Oja's $-y^2w$ normalization \citep{oja1982simplified} with the post-synaptic term clamped to the US) is contiguity learning with homeostasis and has no error term (\textbf{I}).
(iv) Titans \citep{behrouz2024titans} with a linear memory reduces to the delta rule with rate $2\theta_t$ only when momentum and weight decay vanish ($\eta_t=\alpha_t=0$) (\textbf{R}); momentum adds a key-indexed history of signed error to the write, and weight decay adds forgetting, neither of which has a classical counterpart. With an MLP memory, no linear-state correspondence holds.
(v) Gated DeltaNet reduces to Rescorla--Wagner when $\alpha_t\equiv 1$, and adds global multiplicative forgetting otherwise (\textbf{R}).
\end{corollary}

\begin{table}[tp]
\centering\small
\caption{Graded dictionary between linear-attention write rules and classical learning models. \textbf{E} (Exact): the write rule equals the learning model term for term (Theorem~\ref{thm:rw}, Corollary~\ref{cor:dict}). \textbf{R} (Restricted): the rule reduces to an E-tier model only on the stated parameter subset. \textbf{I} (Inspired): the rule borrows a mechanism from the named model without being a special case of it, so predictions do not transfer automatically. Blocking requires an error term: \textbf{[D]} by Proposition~\ref{prop:block} and Appendix~\ref{app:gated_block}; \textbf{[M]} measured in recurrences or trained models (Section~\ref{sec:experiments}).}
\label{tab:dict}
\setlength{\tabcolsep}{4pt}
\resizebox{\textwidth}{!}{%
\begin{tabular}{@{}>{\raggedright\arraybackslash}p{0.21\textwidth}>{\raggedright\arraybackslash}p{0.31\textwidth}cccc>{\raggedright\arraybackslash}p{0.15\textwidth}@{}}
\toprule
Variant & Learning model & Tier & Err. & Forgetting & Cue-sp.\ rate & Blocking \\
\midrule
Linear attn.$^{a}$ & Hebbian contiguity & E & --- & --- & --- & no [D]\\
Decay rule$^{b}$ (RetNet, $\theta{=}0$; GLA, const.\ scalar gate) & Contiguity + exponential forgetting & E & --- & yes & --- & no [D]\\
DeltaNet$^{c}$ & Rescorla--Wagner (1972) & E & yes & --- & --- & \textbf{yes} [D]\\
Gated DeltaNet$^{d}$ & RW when $\alpha_t\equiv1$; RW + global forgetting otherwise & R & yes & yes & --- & yes, shrinks with $p_2$$^{g}$\\
Titans$^{e}$, linear memory & RW when $\eta_t{=}\alpha_t{=}0$; adds momentum \& forgetting otherwise & R & yes & yes & --- & RW limit only$^{g}$\\
Titans$^{e}$ (MLP memory) & no linear-state correspondence & --- & yes & yes & --- & not derived\\
Oja-type$^{h}$ & Oja normalization, post-synaptic clamped to US & I & --- & homeostatic & --- & $\approx$0 [D]\\
\textbf{PH-attention} & RW--Pearce--Hall hybrid$^{i}$ & I & yes & ---$^{f}$ & \textbf{yes} & \textbf{yes} [M, rec.]\\
Softmax attn. & no entry (non-parametric retrieval) & --- & --- & --- & --- & no [M]\\
\bottomrule
\end{tabular}}

\smallskip
{\footnotesize $^{a}$\citet{katharopoulos2020transformers};
$^{b}$\citet{sun2023retentive, yang2024gla};
$^{c}$\citet{schlag2021linear};
$^{d}$\citet{yang2024gated};
$^{e}$\citet{behrouz2024titans}.
$^{f}$PH-attention's memory state $\Sm_t$ does not decay (Eq.~\ref{eq:ph2}); decay acts on the associability trace $\av_t$.
$^{g}$Gated DeltaNet blocking follows from Proposition~\ref{prop:gated_block}; for Titans, only the $\eta_t=\alpha_t=0$ limit is derived.
$^{h}$Trained rate fixed at $0.5$ (Appendix~\ref{app:exp}).
$^{i}$Scales the RW error update by a Pearce--Hall associability trace \citep{li2011differential,tzovara2018human} indexed by key features (Section~\ref{sec:ph}).}
\end{table}

Softmax attention has no write-rule entry in Table~\ref{tab:dict} because it retains all past tokens and retrieves via normalized similarity; we analyze its functional form directly and include it as a primary empirical baseline. Throughout, the correspondence addresses the fast, inference-time formation of in-context associations rather than the slow outer-loop gradient optimization of $\W_Q,\W_K,\W_V$. Appendix~\ref{app:glossary} provides a plain-language glossary mapping every conditioning concept to its attention-layer counterpart (Table~\ref{tab:glossary}) and details what each dictionary tier preserves (Table~\ref{tab:preserved}).

\section{Behavioral Consequences: What Is Derived, What Is Measured}
\label{sec:predictions}

Table~\ref{tab:preds} summarizes the behavioral consequences examined in this paper. Their algebraic components (\textbf{[D]}) establish exact baselines for blocking, backward blocking, extinction, re-pairing persistence, single-state recovery, and readout generalization. Their empirical components (\textbf{[M]}) test whether these distinctions survive end-to-end training on generic in-context association, how capacity scales with dimension and head count, and when a feature-indexed associability state improves task performance.

\begin{table}[tp]
\centering\scriptsize
\caption{Summary of behavioral consequences. \textbf{[D]} follows algebraically from the update rule; \textbf{[M]} is measured in recurrences or trained models. Setting: A, analytic recurrence; T1/T2, trained models with 1 or 2 layers. Analytic (A) and trained (T) blocking indices lie on different scales and are compared by sign, ordering, and within-family trends. Full numerical tables appear in Appendix~\ref{app:glossary} (Tables~\ref{tab:allresults}--\ref{tab:allresults_cont}).}
\label{tab:preds}
\setlength{\tabcolsep}{3pt}
\begin{tabular}{@{}c>{\raggedright\arraybackslash}p{0.15\textwidth}>{\raggedright\arraybackslash}p{0.11\textwidth}>{\raggedright\arraybackslash}p{0.08\textwidth}>{\raggedright\arraybackslash}p{0.34\textwidth}>{\raggedright\arraybackslash}p{0.18\textwidth}@{}}
\toprule
& Consequence & Status & Setting & Headline result & Licensed by \\
\midrule
P1 & Blocking in recurrences & [D] Prop.~\ref{prop:block}, \ref{prop:gated_block} & A & Delta $0.942 = 1-(1-\alpha)^{p_1}$; contiguity $0.000$; gated delta shrinks with $p_2$ & E: RW vs.\ contiguity; R: gated delta \\
P2 & Blocking survives training & [M] & T1, T2 & Summation $0.259$ vs.\ $0.020$ (T1); cosine $0.084$ vs.\ $0.002$; persists up to key cosine $0.55$ & E supplies hypothesis; tested under learned keys \\
P2$'$ & Backward blocking & [D] Prop.~\ref{prop:backward} & A & Delta unchanged ($0.500\to0.500$) or rises; exact least squares (MesaNet) drops ($0.500\to0.000$) & Separates one-step RW from full least squares \\
P3 & Extinction needs error & [D] Prop.~\ref{prop:ext}; [M] & A; T1 & A: delta $97\%$, Hebb $0\%$, decay $18\%$ (passive). T1: delta $13.9\%$, PH $12.3\%$, softmax $94.2\%$, contiguity $\le0.4\%$ & E: RW unlearning vs.\ contiguity \\
P4 & Override persistence & [D] Prop.~\ref{prop:repair}; [M] & A; T1 & A: Hebb $1$, delta $(1-\beta)^{p_2}$. T1: delta $0.065$ vs.\ Hebb $0.689$, softmax $0.676$ & E \\
P5 & Recovery \& renewal & [D] Prop.~\ref{prop:ext}; [M] & A; T1--T2 & A: $\le$\RecoveryMaxRisePct\%. T1--T2: controlled rise $\le$\RecMaxAny\%; softmax shows ABA renewal ($+38.8\%$, $+43.8\%$) & E (erasure) vs.\ token retention (softmax) \\
P6 & Generalization gradient & [D] functional form & A & Linear in $\rho$ (linear attn.); exponential in $\rho$ / Gaussian in distance (softmax, $R^2=0.9998$) & Readout algebra \\
P7 & Capacity of one state & [D] Thm.~\ref{thm:cap}; [M] & A (sphere keys) & Exponents: $1.08 / 1.83$ (all-items), $1.22 / 1.89$ (per-item); collapses to $n^\ast=1$ for uncentered \texttt{elu+1} & Correlation-matrix memory \\
P7$'$ & Partition across heads & [M] & T1 grid & $R^2=0.98$ for log state size $Hd_{\mathrm{head}}^2$; independent replication ($p^H$) rejected & Empirical scaling \\
P8 & PH volatility tracking & [D] Prop.~\ref{prop:vol}; [M] & A; T2 & Rate ratio $1.75$; null at noise $0.35$ (oracle $+0.072$); gain $+0.0033$ (10/10 seeds) at noise $0.1$ & I: RW--PH hybrid \\
\bottomrule
\end{tabular}
\end{table}

\paragraph{Forward and gated blocking.} After cue $A$ is paired with US $U$, pairing the compound $A{+}B$ with $U$ should leave $B$ with minimal associative strength under error-correcting rules, but full strength under contiguity rules.

\begin{proposition}[Closed-form blocking, {[D]}]
\label{prop:block}
Let cues $a,b,c$ be orthonormal, the compound be $a+b$, and consider $p_1$ pairings of the phase-1 cue with $U$ followed by $p_2$ compound pairings. Define the response to $B$ as $\langle b\Sm, U\rangle$ and the blocking index as $1-R_{\mathrm{blocked}}/R_{\mathrm{control}}$, where the control substitutes an irrelevant cue $c$ in phase~1.
(i) For the Hebbian and decay rules, the blocking index is exactly $0$; for the Oja-type rule, it is $0$ to leading order in the homeostatic term (measured \EncSparseOja{} under the encoding of Section~\ref{sec:experiments}).
(ii) For the delta rule with rate $\alpha\in(0,1)$, the blocking index is exactly $1-(1-\alpha)^{p_1}$, independently of $p_2$.
\end{proposition}

Adding a forgetting gate $g\in(0,1]$ (as in Gated DeltaNet) modifies this invariance and separates gated from ungated delta rules:

\begin{proposition}[Gated-delta blocking, {[D]}]
\label{prop:gated_block}
Under the setting of Proposition~\ref{prop:block}, let $\Sm\leftarrow g\Sm+\eta\,k^\top(u-k\,g\Sm)$ with constant gate $g\in(0,1]$ and rate $\eta$. Writing $\rho_1=g(1-\eta)$, $\rho_2=g(1-2\eta)$, and $A=\eta(1-\rho_1^{p_1})/(1-\rho_1)$, the blocking index after $p_2$ compound trials is
\begin{equation}
\text{blocking index}=\frac{\eta\,g\,A\sum_{t=1}^{p_2}g^{p_2-t}\rho_2^{t-1}}{r^{\mathrm{ctl}}_{p_2}} ,
\label{eq:blockgated_main}
\end{equation}
where $r^{\mathrm{ctl}}_{p_2}$ is the control response to $B$. For $g<1$, blocking is positive and decreases monotonically with $p_2$ as $A$'s association decays while $B$ continues learning (Proof and numerical check in Appendix~\ref{app:gated_block}).
\end{proposition}

\paragraph{Backward blocking separates one-step error correction from exact least squares.}
Online associative-optimization frameworks \citep{behrouz2025connected} place DeltaNet and exact in-context least-squares layers such as MesaNet \citep{vonoswald2025mesanet} under a single regression objective, differing only in their optimizer. Conditioning theory provides a crisp protocol—\emph{backward blocking} \citep{dayan2001explaining}—that separates them qualitatively.

\begin{proposition}[Backward blocking, {[D]}]
\label{prop:backward}
Train a compound of cues $A$ and $B$ to a unit outcome $U$ for $n_1$ trials, then $A$ alone to $U$ for $n_2$ trials, and probe $B$.
(i) Under the DeltaNet, Hebbian, and PH recurrences with orthonormal $A$ and $B$, the response to $B$ after phase~2 equals its value after phase~1; with unit keys overlapping by $\rho>0$, it rises.
(ii) Under exact least squares with ridge $\lambda$ (the objective solved by MesaNet at every step), the response to $B$ falls: for an additive orthonormal compound, from $n_1/(\lambda+2n_1)$ after phase~1 to $\lambda n_1/(\lambda^2+\lambda(2n_1+n_2)+n_1n_2)$ after phase~2 (tending to $0$ as $\lambda\to 0$); for a unit-normalized compound with overlap $\rho<1$, from $s/2$ to $s-1$ as $\lambda\to 0$, where $s=\sqrt{2+2\rho}<2$ (Proof and verification in Appendix~\ref{app:backward}).
\end{proposition}

\paragraph{Extinction, override persistence, and spontaneous recovery.}
Non-reinforced presentations $(\kv,\mathbf{0})$ should abolish an association cue-specifically under error correction, decay it passively under trace rules, and leave it intact under contiguity. Similarly, re-pairing a cue $A$ from an initial outcome $U$ to an orthogonal outcome $W$ sharply distinguishes overwriting from superposition. In animals, an extinguished response can recover over time or when returning to the acquisition context, indicating that extinction stores a competing context-gated trace rather than erasing the original \citep{bouton2004context}.

\begin{proposition}[Extinction and single-state non-recovery, {[D]}]
\label{prop:ext}
Under the Hebbian rule, a trial $(\kv,\mathbf{0})$ leaves $\Sm$ unchanged; under the decay rule, the response falls by $\gamma^{p}$ regardless of which cue is presented (passive forgetting); under the delta rule, $\phi(\kv)\Sm\to 0$ geometrically and cue-specifically. Furthermore, let $A$ be extinguished with $R_T=\langle\phi(A)\Sm_T,U\rangle\ge0$. In a single Hebbian, decay, or delta recurrence, if all subsequent write-key features are orthogonal to $\phi(A)$, then $R_t$ remains constant under Hebbian and delta and decays geometrically under decay, precluding recovery.
\end{proposition}

\begin{proposition}[Override persistence under re-pairing, {[D]}]
\label{prop:repair}
Let a unit cue $A$ be paired with $U$ for $p_1$ trials and then with $W\perp U$ for $p_2$ trials, followed by $n$ unrelated items whose keys are orthogonal to $A$. The fraction of the initial association remaining, $\langle A\Sm,U\rangle/R_{\text{acq}}$, is: $1$ under the Hebbian rule; $\gamma^{p_2+n}$ under the decay rule; $(1-\beta)^{p_2}$ under the delta rule with rate $\beta$; $((1-\beta)\gamma)^{p_2}\gamma^n$ under Gated DeltaNet; $(1-\bar b)^{p_2}$ under PH-attention with constant effective rate $\bar b$; and $p_1/(p_1+p_2)$ under a softmax readout over identical keys without positional terms ($n=0$) (Proof and verification in Appendix~\ref{app:override}).
\end{proposition}

Proposition~\ref{prop:ext} applies strictly to single-state recurrences under orthogonal intervening keys because these rules lack a mechanism to store a second, context-dependent trace; Section~\ref{sec:experiments} tests non-orthogonal intervening writes, multilayer stacks, and ABA context-renewal protocols empirically.

\paragraph{Generalization and cue overload.}
For linear attention, response as a function of cue cosine similarity $\rho$ is linear in $\rho$, directly reflecting the inner-product readout. For softmax attention over unit-norm cues, the response takes the form $ae^{\rho/\tau}/(ae^{\rho/\tau}+b)$, which is exponential in $\rho$—equivalently Gaussian in Euclidean distance since $\|a-b\|^2=2-2\rho$—prior to saturation. Note that this differs from Shepard's universal law \citep{shepard1987toward}, which is exponential in distance within a metric psychological space rather than squared Euclidean distance. Finally, cue overload—the degradation of retrieval as stored items accumulate—is described only qualitatively in classical conditioning; Section~\ref{sec:capacity} makes it quantitative along the two architectural axes of a transformer: state dimension and head count.

\section{Cue Overload: The Capacity of an Attention State}
\label{sec:capacity}

The number of associations an attention state can hold depends critically on what the downstream readout requires: the capacity exponent differs by a full power of $d_k$ between faithful value reconstruction and winner-take-all identification. The analytical ingredients extend classical correlation-matrix memory crosstalk \citep{willshaw1969non, kohonen1972correlation} and complement modern Hopfield capacity results \citep{ramsauer2021hopfield} as well as empirical studies of recall--state-size trade-offs \citep{arora2023zoology, jelassi2024repeat}.

\begin{theorem}[Two capacity regimes, {[D]}]
\label{thm:cap}
Let $\Sm=\sum_{j=1}^n \kv_j^\top\vv_j$ with $\kv_j\in\R^{d_k}$, $\vv_j\in\R^{d_v}$ independent and uniform on the unit spheres, and $\mathbf{r}_m=\kv_m\Sm$. Let $\ell=\log(4n^2/\epsilon)$. There is a universal constant $C>0$ such that:
\begin{enumerate}\itemsep2pt
\item \emph{(Fidelity.)} $\E\|\mathbf{r}_m-\vv_m\|^2=(n-1)/d_k$ exactly. Squared relative error at most $\delta$ simultaneously for all $n$ items with probability at least $1-\epsilon$ holds provided
\begin{equation}
\frac{n-1}{d_k}\left[1+C\left(
\sqrt{\frac{\ell}{n}}+\frac{\ell}{n}
+\sqrt{\frac{\ell}{d_v}}+\frac{\ell}{d_v}\right)\right]\le\delta .
\label{eq:fidcond}
\end{equation}
For fixed $\delta,\epsilon$ and $d_v\gtrsim\log d_k$, concentration yields a matching converse up to constants, so $n^\ast = \Theta(\delta d_k)$.
\item \emph{(Identification.)} If the readout need only select the correct $\vv_m$ from the stored set, all $n$ items are identified correctly with probability at least $1-\epsilon$ provided
\begin{equation}
C \ell\left(\frac{1}{d_k}+\frac{1}{d_v}+\frac{n}{d_kd_v}\right)<1.
\label{eq:idcond}
\end{equation}
With $d_v=d_k$ and $n\gg d_k$, this yields $n\log(n/\epsilon)\lesssim d_k^2$, giving $n^\ast=\Omega(d_k^2/\log(d_k/\epsilon))$.
\end{enumerate}
\end{theorem}

Theorem~\ref{thm:cap} establishes tight $\Theta(d_k)$ scaling for fidelity and a near-quadratic achievability bound for identification (verified empirically in Section~\ref{sec:exp_cap}). Consequently, a layer required to reconstruct continuous values faithfully saturates linearly in $d_k$, whereas one that selects from a discrete candidate set can store nearly quadratically many pairs. Crucially, both bounds rely on zero mean pairwise overlap between keys ($\E\langle\kv_i,\kv_j\rangle=0$). Uncentered positive feature maps break this condition: under $\mathrm{normalize}(\mathrm{elu}(g)+1)$ on random Gaussian inputs, mean key overlap sits at $0.685$ for every $d_k$, collapsing Hebbian fidelity capacity to $n^\ast=1$ item (Section~\ref{sec:exp_cap}).

\paragraph{One state, or many?} A transformer layer maintains $H$ heads of width $d_{\text{head}}$ rather than a single state. Whether splitting a state across heads buys capacity or redundancy depends on head specialization. With $p$ bounding a single head's failure probability, redundant heads holding the same association yield a layer failure rate $r\lesssim Lp^H$ if failures are independent, whereas specialized heads carrying distinct required information yield $r\lesssim LHp$ (Appendix~\ref{app:err_remark}). Because these standard union bounds differ by a factor exponential in $H$ and the redundancy bound assumes independent failures across heads sharing a residual stream, Section~\ref{sec:exp_cap} resolves the question empirically across a full $(H,d_{\mathrm{head}})$ grid.

\section{PH-Attention: A Rescorla--Wagner/Pearce--Hall Hybrid}
\label{sec:ph}

Whereas Rescorla--Wagner holds cue salience fixed, \citet{pearce1980model} proposed that a cue's associability—its readiness to enter into new learning—tracks a low-pass-filtered history of unsigned prediction error. Crucially, associability belongs to the stimulus rather than to global trial surprise. We instantiate this stimulus-specific associability for distributed keys via a second feature-indexed state $\av_t\in\R^{d_k}$, reading a cue's associability through its overlap with squared key features:
\begin{align}
\mathbf{e}_t &= \vv_t-\phi(\kv_t)\Sm_{t-1}, \\
\alpha_t &= \frac{\langle \phi(\kv_t)^{\circ 2},\,\av_{t-1}\rangle}
                 {\max(\overline{\av}_{t-1},\epsilon)}, \label{eq:ph_alpha}\\
\beta_t &= \sigma(\xv_t\W_\beta)\cdot 2\sigma(w\alpha_t+b), \label{eq:ph1}\\
\Sm_t &= \Sm_{t-1}+\beta_t\,\phi(\kv_t)^\top\mathbf{e}_t, \label{eq:phS}\\
\av_t &= (1-\eta)\,\av_{t-1}
          +\eta\,\|\mathbf{e}_t\|\,\phi(\kv_t)^{\circ 2},
\label{eq:ph2}
\end{align}
where $\sigma(z)=(1+e^{-z})^{-1}$, $\W_\beta$ is a learned token-to-gate projection, $w$ and $b$ are learned per-head scalars, $\circ 2$ denotes elementwise squaring, $\overline{\av}$ is the feature mean, $\epsilon>0$ is a numerical floor, and $\eta$ is a learned per-head decay rate. Initializing $\av_0=\mathbf{0}$ and $w=b=0$ ensures that PH-attention contains DeltaNet at initialization. For unit-norm keys, normalizing by $\overline{\av}_{t-1}$ renders $\alpha_t$ dimensionless and equal to $1$ under uniform associability (defined as $0$ at $t=0$).

\paragraph{Relation to Pearce--Hall (1980).} The original Pearce--Hall excitatory rule $\Delta V = S\alpha\lambda$ contains no error term, halting learning only when associability $\alpha=|\lambda-\sum V|$ decays to zero. Equation~\ref{eq:phS} retains the Rescorla--Wagner error $\mathbf{e}_t$ and scales its learning rate by $\alpha_t$, forming a hybrid RW--PH rule of the class used to model human conditioning \citep{li2011differential, tzovara2018human}. Here associability is indexed by key features, a single rate per head and token modulates the update, and error magnitude is measured by $\|\mathbf{e}_t\|$.

\paragraph{Relation to existing adaptive gating.} Among recurrent sequence models, none maintains an explicit feature-indexed history of unsigned error to set learning rates. Gated DeltaNet \citep{yang2024gated}, GLA \citep{yang2024gla}, and Mamba-2 \citep{dao2024transformers} compute gates from the current token $\xv_t$; Titans \citep{behrouz2024titans} accumulates signed error momentum into the memory write itself while its gates $\eta_t,\theta_t,\alpha_t$ remain functions of $\xv_t$; RWKV \citep{peng2024eagle} and TTT \citep{zhang2025ttt} likewise lack a cue-indexed unsigned error rate trace; and recursive-least-squares layers maintain feature-indexed gains driven purely by key history rather than error magnitude (hence yielding ratio $1$ below). To isolate feature indexing experimentally, we compare against a \emph{global-associability control} that collapses $\av_t$ in Eq.~\ref{eq:ph2} into a single per-head scalar shared across all cues.

\begin{figure}[t]
\centering
\includegraphics[width=0.92\textwidth]{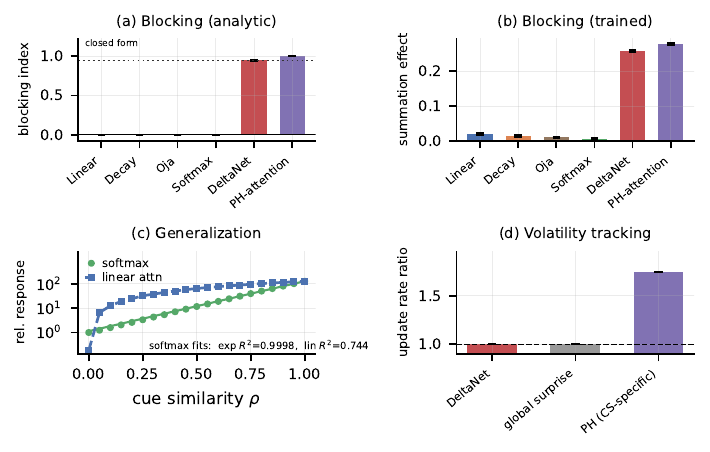}
\caption{\textbf{Conditioning phenomena in attention.} (a) Analytic blocking index under the elementary-feature encoding; the dotted line marks Proposition~\ref{prop:block}'s closed form ($0.942$), while contiguity rules sit at $0.000$. (Softmax is from the dense-encoding run.) (b) In-range summation test ($4+4$ trials, probe at index 12) in one-layer models trained solely on generic in-context association. (c) Readout generalization gradients: prior to saturation, softmax is exponential in similarity $\rho$ (Gaussian in Euclidean distance) while linear attention is linear in $\rho$. (d) With matched base gates, PH-attention's feature-indexed associability trace produces cue-dependent volatility tracking, whereas DeltaNet and the global-associability control (labeled ``global surprise'' in the plot) sit at $1.000$.}
\label{fig:conditioning}
\end{figure}

\begin{proposition}[Volatility tracking, {[D]}]
\label{prop:vol}
Consider two unit-norm, equally familiar cues: one whose outcome has varied across presentations and one whose outcome has been constant. Evaluate counterfactual one-trial updates from the same pre-test state while holding the token-computed base gate $\sigma(\xv_t\W_\beta)$ equal (Figure~\ref{fig:conditioning}d). Under the delta rule and the global-associability control, the update-rate ratio is exactly $1$. Under Eqs.~\ref{eq:ph_alpha}--\ref{eq:ph2}, whenever $\alpha^{\mathrm{vol}}>\alpha^{\mathrm{sta}}$ and $w>0$, the ratio $\sigma(w\alpha^{\mathrm{vol}}+b)/\sigma(w\alpha^{\mathrm{sta}}+b)$ is strictly greater than $1$.
\end{proposition}

\section{Experiments}
\label{sec:experiments}

\paragraph{Protocol.} We evaluate our predictions at two levels. \emph{Analytic experiments} instantiate the parameter-free state recurrences with controlled stimulus vectors to verify algebraic behavior and boundary sensitivity. \emph{Trained experiments} evaluate 1- and 2-layer models ($4$ heads, $d_{\text{model}}=128$, no MLP unless noted) trained solely on generic in-context association and then probed on conditioning protocols never seen during training. Unless stated otherwise, cues use an elementary-feature encoding (disjoint supports of size $k=4$) so compounds are additive sums at the input. Passing through the input projection, pre-LayerNorm, $\W_K$, $\phi=\mathrm{elu}+1$, and $\ell_2$-normalization gives trained DeltaNet cue keys a baseline overlap of $\cos(k_A,k_C)\approx 0.22$ and partial compound additivity (cosine $0.85$ with the normalized element sum; Appendix~\ref{app:extra}). All error bars denote $\pm 1$ s.e.m.\ over 3--5 model seeds (10 seeds where stated). Full experimental details and comprehensive tables appear in Appendices~\ref{app:exp} and \ref{app:extra}.

\subsection{Blocking and the Theory-to-Trained Bridge (P1, P2)}

\paragraph{Analytic recurrences [D].} Under Proposition~\ref{prop:block}'s elementary-feature encoding ($\alpha=0.3, p_1=p_2=8$), contiguity rules yield blocking indices of \EncSparseHebb{} (Hebbian), \EncSparseDecay{} (decay), and \EncSparseOja{} (Oja), while DeltaNet yields \EncSparseDelta{} and PH-attention \EncSparsePh{} (Figure~\ref{fig:conditioning}a). Sweeping $\alpha$ across \DeltaAlphaN{} values matches $1-(1-\alpha)^{p_1}$ to float32 rounding (maximum error \DeltaAlphaMaxErr). Violating the additive-compound assumption by renormalizing a dense compound lowers DeltaNet blocking to \EncDenseNormDelta{} (matching Eq.~\ref{eq:blocknorm}'s $0.6663$). Across an analytic sweep of cue overlap $\rho\in[0,0.6]$ and configural compound weights $\eta\in[0,0.5]$ (Appendix Figure~\ref{fig:sweep}), blocking decreases smoothly but remains positive in every cell for error-correcting rules (down to $0.098$ at $\rho=0.6,\eta=0.5$), while contiguity rules remain at zero.

\paragraph{Trained models across two readouts [M].} Because cosine similarity to the US is scale-invariant, it can saturate when a weak association still points toward $U$. We therefore measure trained blocking using both a direct cosine index and a magnitude-free \emph{summation test} that probes the compound $B{+}G$, where $G$ was independently paired with an orthogonal outcome $W$: a blocked $B$ steers the output toward $W$, whereas an unblocked $B$ produces a $U/W$ mixture. In one-layer models (base mean cosine \IclDeltaLone{} for DeltaNet, \IclHebbLone{} for Hebbian at 8 pairs), the direct cosine index gives \CosDeltaLone{} (DeltaNet) and \CosPhLone{} (PH) against \CosHebbLone{} (Hebbian), \CosDecayLone{} (decay), and \CosOjaLone{} (Oja). The summation test (4 phase-1 and 4 phase-2 trials, probe at index 12 inside the trained positional range) yields \SumDeltaLone{} (DeltaNet) and \SumPhLone{} (PH) against \SumHebbLone{} (Hebbian), \SumDecayLone{} (decay), \SumOjaLone{} (Oja), and \SumSoftmaxLone{} (softmax), with a paired DeltaNet$-$Hebbian difference of \SumDeltaHebbLoneCI{} (Figure~\ref{fig:conditioning}b; Appendix Table~\ref{tab:sumdesigns}). Two-layer models replicate this ordering (\SumDeltaSP{} vs.\ \SumHebbSP{}; paired difference $+0.232$ $[+0.031, +0.433]$).

Moreover, Proposition~\ref{prop:block} predicts that blocking rises across $p_1\in\{1,2,4\}$ phase-1 trials before saturating near $p_1=4$ at trained write rates ($\approx 0.68$). Matching probe positions with disjoint leading filler pairs confirms this directional dependence across all four trained model families: one-layer DeltaNet rises by \SumDoseDeltaLone{} from $p_1=1$ to $p_1=4$ (relative ratios \SumDoseRatioOneLone{} and \SumDoseRatioTwoLone{} of the 4-trial value), whereas contiguity rules and softmax shift by at most $0.015$.

\paragraph{Mechanistic decomposition and dense-cue scaling.} To relate Proposition~\ref{prop:block}'s closed form ($0.96$ at trained head rates) to the trained cosine index ($0.083$), we replay each head's recurrence on its captured keys, values, and gates (Table~\ref{tab:ladder}). Replay reproduces trained head outputs exactly, averaging $0.466$ across heads ($0.515$ pooled). One-at-a-time substitution reveals that unit-norm, non-additive compound keys account for nearly the entire drop from $0.96$ to $0.47$, while the remaining attenuation to $0.143$ (output magnitude) and $0.083$ (cosine index) arises in the output projection, final layer norm, and cosine normalization. Similarly, a $2^3$ factorial across stimulus dimension $d_{\text{stim}}\in\{32,128\}$, depth, and compound training ($48$ models; Appendix Table~\ref{tab:factorial}) demonstrates why cosine blocking appears suppressed ($0.010$) in two-layer $d_{\text{stim}}=32$ dense models: at $d_{\text{stim}}=32$, the control cosine sits near ceiling ($0.96$--$0.99$), masking blocking ($r=-0.90$ across all 24 DeltaNet models), whereas head-level blocking ($0.365$ $[0.296, 0.434]$) and summation blocking ($+0.209$ $[+0.154, +0.264]$) remain strong. Finally, explicitly raising test-cue overlap $\rho$ up to $0.6$ (raising trained key cosine from $0.22$ to $0.55$; Appendix Table~\ref{tab:overlap}) attenuates but never eliminates blocking: head-level blocking falls from $0.47$ to $0.28$, cosine from $0.084$ to $0.025$, and summation from $0.259$ to $0.170$ (DeltaNet$-$Hebbian $+0.158$ $[+0.139,+0.177]$).

\begin{table}[t]
\centering\scriptsize
\caption{Mechanistic gap ladder from Proposition~\ref{prop:block} to the trained cosine blocking index (12-head mean across 3 models).}
\label{tab:ladder}
\setlength{\tabcolsep}{4pt}
\begin{tabular}{@{}>{\raggedright\arraybackslash}p{0.62\textwidth}cc@{}}
\toprule
Level & delta & PH \\
\midrule
(a) Proposition~\ref{prop:block}, $\alpha=0.3, p_1=8$ & 0.942 & 0.942 \\
(b) Closed form at each head's phase-1 write rate & 0.956 & 0.949 \\
(b$'$) Unit-compound closed form (Eq.~\ref{eq:blocknorm}) at measured rate & 0.676 & 0.671 \\
Replay: orthonormal additive cues, constant rate \& value & 0.957 & 0.938 \\
\quad + actual per-token write rates & 0.958 & 0.976 \\
\quad + actual per-token values & 1.160 & 1.206 \\
\quad + renormalized compound & 0.817 & 0.852 \\
\quad + actual element keys, compound $k_A+k_B$ & 0.778 & 0.830 \\
\quad + compound $\mathrm{norm}(k_A+k_B)$ & 0.553 & 0.574 \\
\quad + actual compound key & 0.427 & 0.438 \\
\quad + actual probe query & 0.438 & 0.451 \\
(c) Exact head replay = trained head readout (pooled) & 0.466 (0.515) & 0.468 (0.520) \\
(d$'$) Model output magnitude index & 0.143 & 0.163 \\
(d) Model output cosine index & 0.083 & 0.092 \\
\bottomrule
\end{tabular}
\end{table}

\subsection{Extinction, Override Persistence, Recovery, and Renewal (P3--P5)}
\label{sec:exp_ext}

\begin{figure}[t]
\centering
\includegraphics[width=0.85\textwidth]{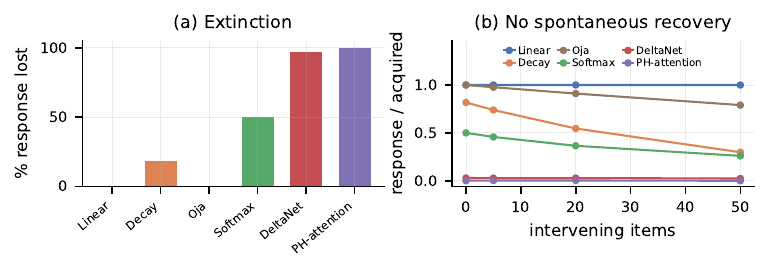}
\caption{\textbf{Extinction and single-state non-recovery in analytic recurrences.} (a) Percentage of acquired response lost after non-reinforced trials: Hebbian and Oja cannot extinguish; decay loses response passively; softmax's 50\% reflects uniform averaging over equal numbers of reinforced and non-reinforced tokens; DeltaNet and PH-attention extinguish via error correction. (b) Response across intervening retention items: no recurrence recovers.}
\label{fig:extinction}
\end{figure}

\paragraph{Extinction [D, M].} In analytic recurrences, non-reinforced trials abolish \ExtDelta\% of the acquired response under DeltaNet and \ExtPH\% under PH-attention, compared with \ExtHebb\% (Hebbian), \ExtOja\% (Oja), \ExtDecay\% (decay, purely passive), and \ExtSoftmax\% (softmax, reflecting uniform averaging over equal trial counts; Figure~\ref{fig:extinction}a). In trained one-layer models probed inside the trained positional range (8 acquisition + 8 extinction trials, probe at index 16), DeltaNet loses \TrExtDeltaLone\% (\TrExtDeltaLoneFill\% when filler pairs replace extinction trials) and PH-attention loses \TrExtPhLone\%, against at most $0.4\%$ for contiguity rules (\TrExtHebbLone\% for Hebbian) and \TrExtSoftmaxLone\% for softmax. Where extinction occurs in trained DeltaNet, it is cue-specific: in a two-cue protocol, the extinguished cue drops by $10.5\pm 0.7\%$ versus $1.6\pm 0.1\%$ for the non-extinguished cue.

\paragraph{Override persistence under re-pairing [D, M].} Re-pairing cue $A$ from outcome $U$ to an orthogonal outcome $W$ sharply separates overwriting from superposition (Proposition~\ref{prop:repair}). In analytic recurrences ($p=8$ trials per phase), DeltaNet leaves $0.058$ of the original $U$ association and PH leaves $0.0007$, whereas Hebbian retains $1.000$, decay $0.851$, Oja $0.982$, and softmax $0.500$ (Table~\ref{tab:override}). In trained one-layer models ($5+5$ trials, measured relative to $W$-only and no-override controls), the persistent fraction of the original association is $0.065\pm 0.003$ for DeltaNet and $0.356\pm 0.003$ for PH, compared with $0.689\pm 0.001$ (Hebbian), $0.668\pm 0.001$ (decay), $0.676\pm 0.001$ (Oja), and $0.676\pm 0.002$ (softmax; paired DeltaNet$-$Hebbian difference $-0.624$ $[-0.636, -0.611]$). Even though softmax extinguishes strongly under zero-outcome trials, it retains re-paired associations as an equal-weight mixture over stored tokens.

\begin{table}[t]
\centering\scriptsize
\caption{Fraction of the original association retained after re-pairing in analytic recurrences (Proposition~\ref{prop:repair}).}
\label{tab:override}
\begin{tabular}{@{}lcccc@{}}
\toprule
Rule & $p=4$ & $p=8$ & $p=8$, $n=50$ & Closed form \\
\midrule
Hebbian & 1.000 & 1.000 & 1.000 & $1$ \\
Decay & 0.922 & 0.851 & 0.310 & $\gamma^{p+n}$ \\
Oja-type & 0.991 & 0.982 & 0.873 & $\approx 1-p\alpha/d$ \\
Softmax & 0.500 & 0.500 & 0.232 & $p_1/(p_1{+}p_2)$ \\
DeltaNet & 0.240 & 0.058 & 0.058 & $(1-\beta)^p$ \\
Gated delta & 0.221 & 0.049 & 0.018 & $((1-\beta)\gamma)^p\gamma^n$ \\
PH-attention & 0.026 & 0.0007 & 0.0007 & $(1-\bar b)^p$ \\
\bottomrule
\end{tabular}
\end{table}

\paragraph{Spontaneous recovery and ABA context renewal [M].}
In the analytic retention experiment (dense cues), no rule recovers (largest rise \RecoveryMaxRisePct\%; Figure~\ref{fig:extinction}b). Evaluating trained models requires two essential experimental controls: (1) probing within the trained positional range ($\le 19$, since unvisited positional embeddings remain at initialization), and (2) subtracting a never-presented-cue baseline, as models exhibit a decaying global bias toward the in-context US. Without these controls, two-layer DeltaNet exhibits an apparent recovery of \RecUncontrolledDelta\% (up to \RecUncontrolledDeltaMax\%); with both controls in place across \RecNmodels{} models and \RecNsettings{} settings, the largest seed-averaged rise drops to \RecMaxAny\% (Appendix Table~\ref{tab:recovery} and Figure~\ref{fig:recovery}). An apparent rise in paired difference (\texttt{extinguished $-$ novel}, up to \RecMaxSpecific\%) reflects faster decay of the novel-cue bias rather than return of the extinguished association.

Across formal equivalence tests (5\% margin), 20 of 21 settings pass (5 of 6 settings with $\ge 10\%$ extinction pass; Appendix~\ref{app:extra}). Substituting blank tokens or zero-outcome cues for random retention fillers eliminates outcome-driven suppression and yields equivalence in all one-layer elementary-feature models (6 of 12 two-layer elementary-feature tests exceed the 5\% bound, and post-hoc blank retention after strong compound extinction produces $+12.4\pm 2.7\%$ above control across 6 one-layer dense DeltaNet models). Most revealingly, an \emph{ABA context-renewal protocol} (acquisition with context token $X_1$, extinction with $X_2$, probe in $X_1$) yields a stark mechanistic dissociation: error-correcting state rules show no renewal beyond cue-plus-context summation in 18 of 21 models (the three exceptions occur in the most strongly extinguishing seeds), whereas softmax attention exhibits strong cue-specific renewal beyond summation ($+38.8\pm 0.9\%$ in one-layer and $+43.8\pm 0.9\%$ in two-layer elementary-feature models; $+14.6$ to $+32.2$ under unnormalized dense readouts), directly reflecting its retention of original acquisition tokens.

\subsection{Generalization Gradients (P6)}
Evaluating generalization across similarity $\rho\in[0,1]$ confirms the readout functional forms (Figure~\ref{fig:conditioning}c). Nonlinear least-squares fits on the response scale favor the exponential-in-$\rho$ model over a linear fit in \GenExpWins{} competition regimes for softmax ($R^2=\GenSoftExpR$ vs.\ \GenSoftLinR{} at \GenSoftRange$\times$ dynamic range), failing only in the low-temperature saturated regime where the logistic flattens ($R^2=\GenSatExpR$ vs.\ \GenSatLinR{} at \GenSatRange$\times$), while linear attention is exactly linear ($R^2=\GenLinR$).

\subsection{Capacity Regimes and Head Partitioning (P7, \texorpdfstring{P7$'$}{P7'})}
\label{sec:exp_cap}

\begin{figure}[t]
\centering
\includegraphics[width=\textwidth]{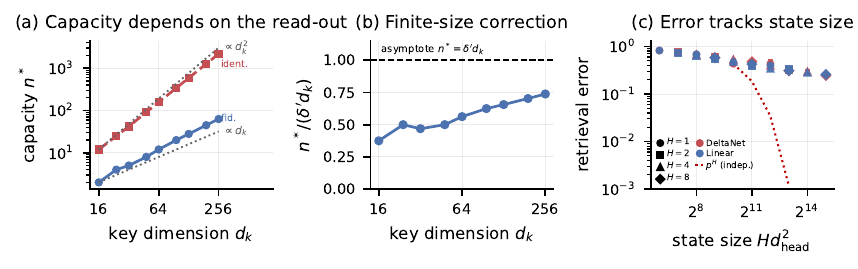}
\caption{\textbf{Capacity of a single attention state and multi-head partitioning.} (a) Largest load $n^\ast$ at 95\% success versus key dimension $d_k$ for Hebbian memory under fidelity and identification criteria (per-item criterion), compared with linear and quadratic guide slopes. (b) Ratio of measured fidelity capacity to asymptotic prediction $\delta' d_k$, approaching $1$ from below as predicted by Eq.~\ref{eq:fidcond}. (c) Retrieval error versus total state size $Hd_{\mathrm{head}}^2$ across the full $H\times d_{\mathrm{head}}$ grid; state size explains $R^2=0.98$ of variance, whereas the independent-replication curve $p^H$ diverges sharply.}
\label{fig:capacity}
\end{figure}

\paragraph{Fidelity versus identification scaling [D, M].} Sweeping $d_k\in[16,256]$ for random unit-sphere keys and values without a feature map confirms the predicted power-of-$d_k$ gap (Figure~\ref{fig:capacity}a). Under the per-item 95th-percentile criterion ($\delta=0.25$ after per-item rescaling), fitted Hebbian exponents are \CapFidHebb{} (fidelity) and \CapIdeHebb{} (identification). Under Theorem~\ref{thm:cap}'s exact all-items, unrescaled criterion ($4{,}000$ independent memories per $d_k$), the exponents sharpen to $1.08$ $[1.06, 1.08]$ (fidelity) and $1.83$ $[1.82, 1.86]$ (identification; $0.92$--$1.32$ and $1.75$--$1.94$ across $\delta,\epsilon$ sweeps; Appendix Table~\ref{tab:cap}). In fidelity, $n^\ast/(\delta' d_k)$ rises toward the asymptote from below (\CapRatioLo{} at $d_k=16$ to \CapRatioHi{} at $d_k=256$; theorem criterion $0.38 \to 0.52 \to 0.69$ across $d_k=64,256,1024$), matching the shrinking finite-size correction in Eq.~\ref{eq:fidcond} (Figure~\ref{fig:capacity}b). Identification exponents differ by rule (Hebbian \CapIdeHebb{}, DeltaNet \CapIdeDelta{}, decay \CapIdeDecay{}, as decay saturates near its timescale $1/(1-\gamma)=50$), and measured SNR matches $d_k/(n-1)$ to \SnrRelErrPct\% relative error.

\paragraph{Collapse under uncentered positive feature maps [M].} Applying $\mathrm{normalize}(\mathrm{elu}(g)+1)$ to Gaussian keys produces a constant mean pairwise cosine of $0.685$ across all $d_k$, violating Theorem~\ref{thm:cap}'s zero-mean overlap condition. Consequently, Hebbian fidelity capacity collapses to $n^\ast=1$ item at every $d_k\le 256$ and identification capacity grows only from $2$ to $5$. Mean-centering the feature map restores zero mean overlap and recovers the spherical scaling laws (exponents within $0.08$).

\paragraph{Total state size governs multi-head error [M].} Sweeping the full grid $H\in\{1,2,4,8\}\times d_{\mathrm{head}}\in\{8,16,32,64\}$ (3 seeds per cell, load 24) decouples total state size $Hd_{\mathrm{head}}^2$ from partitioning across heads. Across the grid, $\log(Hd_{\mathrm{head}}^2)$ accounts for $R^2=\GridRtwoStateDelta$ (DeltaNet) and \GridRtwoStateHebb{} (Hebbian) of the variance in $\log$ retrieval error, compared with \GridRtwoHeadsDelta{}/\GridRtwoHeadsHebb{} for $\log H$ and \GridRtwoWidthDelta{}/\GridRtwoWidthHebb{} for $\log d_{\mathrm{head}}$ (Figure~\ref{fig:capacity}c). At matched state sizes, more heads outperform wider heads in only \GridMoreWinsDelta{} (DeltaNet) and \GridMoreWinsHebb{} (Hebbian) cases (maximum spread \GridSpreadMaxDelta{}). Below single-head capacity (6 pairs), Hebbian error moves from \EasyHeadErrOneHebb{} ($H=1$) to \EasyHeadErrMaxHebb{} ($H=8$), far from the independent-replication prediction \EasyHeadPredHebb{}. Head-ablation diagnostics in 4-head DeltaNet models confirm specialization directly: ablating any active head increases error by $+0.05$--$0.06$, per-item head contributions are uncorrelated ($|r|\le 0.03$), and head outputs have low overlap (cosine $0.09$--$0.22$). At matched parameter count ($H=4,d_{\mathrm{head}}=16$ vs.\ $H=1,d_{\mathrm{head}}=64$), error tracks state size ($0.363\to 0.227$), while at equal state size ($1{,}024$) one head slightly outperforms four ($0.304$ vs.\ $0.363$).

\subsection{Volatility Tracking and Limits of Unsigned Error Gating (P8)}
\label{sec:exp_ph}

\paragraph{Predicted rate signature [D, M].} Under the controlled volatility assay (Proposition~\ref{prop:vol}, $w=0.75$), PH-attention achieves an associability ratio of \VolRatioPhcueAssoc{} and an update-rate ratio of \VolRatioPhcue{} between volatile and stable cues, whereas DeltaNet and the global-associability control remain at exactly \VolRatioDelta{} (Figure~\ref{fig:conditioning}d). Across \VolWN{} positive values of $w$, the rate ratio stays strictly above $1$ ($[\VolWLo,\VolWHi]$).

\paragraph{When unsigned error fails—and when it works [M].} On a regime-switch task where all cues shift simultaneously (offering no per-cue reliability differences), PH-attention matches DeltaNet (\SwPhClean{} vs.\ \SwDeltaClean{} clean; Table~\ref{tab:ph}). We therefore evaluate a \emph{mixed-reliability task} in which half the cues are volatile (fresh outcome each trial) and half are stable (fixed clean outcome observed through additive Gaussian noise of std $0.35$). Across 10 paired seeds, PH-attention yields \VolTaskPh{} against \VolTaskDelta{} for DeltaNet and \VolTaskPhglobal{} for the global-associability control (paired difference \PHPairedDiff{}, $t=\PHPairedT$, 95\% CI $[\PHPairedCIlo,\PHPairedCIhi]$).

\begin{table}[t]
\centering\small
\caption{In-context recall under distribution shift (cosine accuracy, mean $\pm$ s.e.m.\ over 3 seeds for switch, 10 for mixed reliability).}
\label{tab:ph}
\begin{tabular}{@{}lccc@{}}
\toprule
Write rule & Switch (clean) & Switch (noisy) & Mixed reliability \\
\midrule
Linear (Hebb) & \SwHebbClean & \SwHebbNoisy & \VolTaskHebb \\
Decay (RetNet) & \SwDecayClean & \SwDecayNoisy & \VolTaskDecay \\
DeltaNet (RW) & \SwDeltaClean & \SwDeltaNoisy & \VolTaskDelta \\
Global associability (control) & \SwPhglobalClean & \SwPhglobalNoisy & \VolTaskPhglobal \\
\textbf{PH-attention} & \SwPhClean & \SwPhNoisy & \VolTaskPh \\
\midrule
Softmax (full attention) & \SwSoftmaxClean & \SwSoftmaxNoisy & \VolTaskSoftmax \\
\bottomrule
\end{tabular}
\end{table}

To diagnose whether this null reflects task irrelevance, optimization failure, or signal limitation, we evaluate an oracle rate (setting $\beta$ per cue from its true volatile/stable identity) alongside constrained and fixed PH variants (Appendix Table~\ref{tab:p6}). The oracle gains $+0.0718$ $[+0.0682, +0.0755]$ over DeltaNet, proving the task strongly rewards cue-specific rates. Yet even a positive-constrained slope ($w>0$) gains only $+0.0002$ $[-0.0014, +0.0017]$, and fixed positive gates hurt ($-0.016$ to $-0.031$). Inspecting the learned associability trace reveals why: in every trained PH variant at noise $0.35$, the volatile/stable associability ratio never exceeds $1.013$ (vs.\ $2.24$ in the analytic assay). An idealized grid search over one-hot recurrences explains the bottleneck: at noise $0.35$, unsigned prediction error confounds aleatoric outcome noise on stable cues with true outcome volatility, capping any hand-set PH gate at $+0.0013$ over the best single rate (vs.\ $+0.096$ for per-group oracle rates).

We verify this diagnosis on a lower-noise variant (stable-cue noise lowered from $0.35$ to $0.10$) paired with a slower associability trace ($\eta_0=\sigma(-3)=0.047$; Appendix Table~\ref{tab:p6b}). In this regime, learned PH-attention beats DeltaNet in 10 of 10 paired seeds by $+0.0033$ $[+0.0025, +0.0042]$ ($14.3\%$ of the oracle's $+0.0232$ gain), learns a positive slope $w=0.161$ $[0.119, 0.204]$ across $86\%$ of heads, and separates the two cue groups (associability ratio $1.134$). In contrast, on generic one-layer in-context association where cues are stationary, the unconstrained slope drifts slightly negative ($w=\PhLearnedWSeed$ across \PhLearnedWSeedN{} seeds, \PhLearnedWPos{} of \PhLearnedWN{} heads positive), resembling Mackintosh-like down-weighting of noisier cues \citep{mackintosh1975theory}.

\section{Related Work}
\label{sec:related}

\paragraph{Attention as associative memory and online optimization.} Viewing attention as associative memory spans correlation-matrix memories \citep{willshaw1969non, kohonen1972correlation, anderson1972simple}, classical and modern Hopfield networks \citep{hopfield1982neural, ramsauer2021hopfield}, sparse distributed memory \citep{bricken2021attention}, and fast-weight linear transformers \citep{schmidhuber1992learning, ba2016fastweights, schlag2021linear, yang2024parallelizing, yang2024gated}. \citet{behrouz2025connected} unify many recurrent sequence layers as online optimization of an associative objective. While every derived algebraic property in our work can also be framed within online optimization, our conditioning dictionary is orthogonal and complementary: it converts write rules into discriminating behavioral interventions—such as backward blocking (Proposition~\ref{prop:backward}), which separates one-step DeltaNet updates from exact in-context least squares (MesaNet; \citealp{vonoswald2025mesanet}) despite their shared regression objective—and tests which distinctions survive end-to-end training.

\paragraph{In-context learning and neuroscience.} Prior work showing that transformers can implement gradient descent in context \citep{vonoswald2023transformers, akyurek2023what, dai2023why, garg2022what} aligns naturally with Theorem~\ref{thm:rw}, since the delta rule is one step of least-squares descent; our focus is the resulting behavioral phenomenology rather than optimizer identification. Neuroscience mappings linking transformers to hippocampal codes and biologically plausible key--value memories \citep{whittington2022relating, tyulmankov2021biological, zador2023catalyzing} relate representations, whereas we map update rules to behavioral signatures.

\paragraph{Adaptive surprise gating.} Recent recurrent architectures such as Titans \citep{behrouz2024titans} incorporate signed-error momentum into memory writes with token-dependent gates, while Gated DeltaNet \citep{yang2024gated} and related state-space layers gate writes and decay from the current token. PH-attention addresses a narrower question: isolating an explicit feature-indexed unsigned-error associability state that modulates the learning rate, evaluated against a shared scalar global-associability control.

\section{Discussion: Behavioral Diagnostics for Linear and Hybrid Sequence Models}
\label{sec:discussion}

Production-scale sequence models increasingly hybridize softmax attention with linear-attention layers that sit on opposite sides of Table~\ref{tab:dict}'s error-term boundary: Gated DeltaNet in Qwen3-Next \citep{qwen2025qwen3next} and Kimi Delta Attention in Kimi Linear \citep{kimi2025linear} use error-correcting writes, whereas lightning attention in MiniMax-01 \citep{minimax2025minimax01} uses gated contiguity without an error term. Viewing these layers through conditioning theory yields three falsifiable, layer-level diagnostic predictions testable solely via forward passes:

\paragraph{1. In-context belief revision and override persistence.} When an in-context cue is re-paired with a new outcome, error-correcting delta layers overwrite the prior association exponentially by $(1-\beta)$ per trial (retaining $0.065$ in trained one-layer DeltaNet), whereas contiguity layers retain it at full strength and return a frequency-weighted mixture ($0.668$--$0.689$), and softmax layers retain all past tokens ($0.676$). Exchanging an error-correcting linear layer for a contiguity layer therefore alters whether late-context corrections override earlier associations or blend with them.

\paragraph{2. Erasure versus context-dependent retrieval.} Conditioning distinguishes genuine erasure from temporary suppression via extinction, savings, and context renewal. Single-state delta recurrences erase cue-specifically, show no savings (\SavingsDelta$\times$ a fresh cue), and cannot recover under orthogonal writes (Proposition~\ref{prop:ext}). By contrast, softmax attention preserves original tokens in full and therefore exhibits strong context-dependent ABA renewal ($+38.8\%$ to $+43.8\%$ beyond cue-plus-context summation). Hybrid architectures mixing softmax and linear layers may therefore exhibit renewal or recovery driven by their remaining softmax heads.

\paragraph{3. Interference versus temporal decay under context load.} An early in-context association competes with subsequent tokens differently across write rules: in pure Hebbian or delta memory, degradation is driven by key-overlap interference (scaling as $(n-1)/d_k$ for random unit keys and governed by total state size $Hd_{\mathrm{head}}^2$ across heads), whereas decay-based layers lose early items geometrically with elapsed steps regardless of interference. Testing whether long-context recall degrades with token distance or with interfering writes directly diagnoses which mechanism dominates in a given layer.

\section{Limitations}
\label{sec:limits}

Five limitations bound the scope of our conclusions. First, softmax attention has no write-rule entry in Table~\ref{tab:dict} and outperforms all linear rules across our recall benchmarks (Table~\ref{tab:ph}); extending this behavioral analysis to normalized non-parametric retrieval remains an open direction. Second, closed-form predictions such as Proposition~\ref{prop:block} depend on cue geometry: trained keys are non-orthogonal ($\cos\approx 0.22$) and only partly additive ($0.85$), shrinking trained blocking to roughly half the analytic value at the head level and smaller at the output, so the dictionary predicts sign, ordering, and directional trial-count dependence rather than exact trained magnitudes. Third, PH-attention contains free hyperparameters and separates cues only when their squared-key features differ (spanning $[\PhSensBlockLo,\PhSensBlockHi]$ blocking across \PhSensN{} settings). Fourth, our trained experiments use small-scale synthetic architectures ($\le 2$ layers, $d_{\text{model}}=128$, no MLP) and probe fast inference-time memory rather than multi-layer reasoning in large pretrained models. Finally, the layer-level diagnostics discussed in Section~\ref{sec:discussion} have not yet been evaluated in full-scale hybrid language models, where softmax layers, MLPs, and pretraining mixtures may mask or compose with single-layer dynamics.

\section{Conclusion}

Treating linear attention as a named associative-learning rule turns the qualitative ``attention as associative memory'' metaphor into concrete, falsifiable predictions. Our graded dictionary separates algebraic invariants from trained-network behavior, demonstrates that cue competition and belief overwriting survive generic in-context training, reveals sharp divergences between single-state erasure and softmax context renewal, and quantifies how retrieval criteria and feature-map centering govern memory capacity. At the same time, PH-attention illustrates both the promise and discipline of theory-driven architecture design: translating Pearce--Hall associability yields the exact cue-specific rate signature predicted, while controlled experiments expose precisely why unsigned error struggles under aleatoric noise and succeeds only when outcome noise is low. Behavioral paradigms from classical conditioning offer a compact, mechanism-grounded unit-test suite for understanding and diagnosing fast in-context memory in modern sequence architectures.

\appendix

\section{Glossary and Consolidated Results}
\label{app:glossary}

Table~\ref{tab:glossary} defines each conditioning term used in the main text alongside its counterpart in an attention layer. Tables~\ref{tab:allresults} and \ref{tab:allresults_cont} collect the numerical results for every consequence in Table~\ref{tab:preds}. Table~\ref{tab:preserved} summarizes what each entry of Table~\ref{tab:dict} preserves from its classical learning counterpart and what it leaves outside the correspondence. Throughout, \textbf{A} denotes analytic recurrences (parameter-free state updates), and \textbf{T1/T2} denote trained 1- or 2-layer models ($4$ heads, $d_{\text{model}}=128$, no MLP). Error bars ($\pm$) report s.e.m.\ over model seeds. Analytic and trained blocking indices operate on different scales and are compared by sign, ordering, and within-family relative changes rather than raw magnitude.

\begin{table}[tp]
\centering\scriptsize
\caption{Glossary of conditioning terms and their attention-layer counterparts.}
\label{tab:glossary}
\setlength{\tabcolsep}{4pt}
\begin{tabular}{@{}>{\raggedright\arraybackslash}p{0.16\textwidth}>{\raggedright\arraybackslash}p{0.40\textwidth}>{\raggedright\arraybackslash}p{0.40\textwidth}@{}}
\toprule
Term & Plain definition & Counterpart in this paper \\
\midrule
Conditioned stimulus (CS) & A cue (e.g., a tone) that initially elicits no response and comes to predict an outcome. & The key $\phi(\kv_t)$ of a study token (the cue slot of a trained model's input). \\
Unconditioned stimulus (US) & The outcome (e.g., food) that the cue comes to predict. & The value $\vv_t$ paired with the cue (the outcome slot). \\
Contiguity & Learning from co-occurrence alone: every present cue gains strength regardless of existing predictions. & Unnormalized linear attention, $\Sm\leftarrow\Sm+\phi(\kv)^\top\vv$, as well as decay and Oja-type variants (no error term). \\
Cue competition & Co-present cues share a single prediction error, so each cue's learning depends on what others already predict. & Error-correcting write rules (DeltaNet, Gated DeltaNet, PH-attention): the update uses the state's summed prediction over all active features. \\
Blocking \citep{kamin1969predictability} & After $A\to U$, compound training $AB\to U$ teaches $B$ little because $A$ already predicts $U$. & P1/P2. Blocking index $=1-R_{\text{blocked}}/R_{\text{control}}$ on probe $B$ (Prop.~\ref{prop:block}). Trained models: cosine and summation readouts (Section~\ref{sec:experiments}). \\
Backward blocking & After $AB\to U$, solo training $A\to U$ retroactively reduces responding to $B$. & P2$'$ (Prop.~\ref{prop:backward}). Absent in one-step delta rules under orthogonal cues; present under exact least squares (MesaNet). \\
Overshadowing & When two cues are trained together, the weaker acquires less strength than if trained alone. & Relative to an element-alone control, only error-correcting rules overshadow in recurrences (delta $0.135$, PH $0.273$; contiguity $0.990$; softmax $1.000$; Appendix~\ref{app:extra}). \\
Extinction & Presenting a cue repeatedly without the outcome weakens responding. & Writing $(\kv_A,\mathbf{0})$. Delta drives $\phi(A)\Sm\to 0$ cue-specifically; Hebbian is unchanged; decay fades passively (Prop.~\ref{prop:ext}). \\
Spontaneous recovery & An extinguished response returns after a delay, indicating suppression rather than erasure. & Probing $A$ after intervening tokens (P5). In sequence models, elapsed time corresponds to intervening tokens that also alter input history. \\
Context renewal & An extinguished response returns when tested outside the extinction context. & ABA design: acquisition with context token $X_1$, extinction with $X_2$, probe in $X_1$ (Section~\ref{sec:exp_ext}, Appendix~\ref{app:extra}). \\
Savings & Re-learning after extinction is faster than learning a fresh cue. & One re-acquisition trial after extinction relative to a fresh cue: delta $1.06\times$ (no savings); Hebbian and decay never lost the trace (Appendix~\ref{app:extra}). \\
Associability & How readily a cue enters into new learning (its cue-specific learning rate). & Effective write rate $\beta_t$. PH-attention's feature-indexed trace $\av_t$ makes it cue-specific (Eqs.~\ref{eq:ph_alpha}--\ref{eq:ph2}). \\
Generalization gradient & Response to a test cue as a function of similarity to the trained cue. & Response versus cosine similarity $\rho$: linear in $\rho$ for linear attention; exponential in $\rho$ (Gaussian in Euclidean distance) for softmax prior to saturation. \\
\midrule
Associative strength $V$ & Learned strength of a cue--outcome link. & $\langle\phi(a)\Sm,u\rangle$. \\
Salience $\alpha$ & How strongly a cue drives updates. & Key norm and write gate. \\
Compound & Two or more cues presented simultaneously. & Sum of elementary cue vectors at input. \\
Summation test & Probing a cue by combining it with a separately trained cue. & The $B{+}G$ probe in Section~\ref{sec:experiments}. \\
\bottomrule
\end{tabular}
\end{table}

\begin{table}[tp]
\centering\scriptsize
\caption{Consolidated results across all behavioral consequences in Table~\ref{tab:preds}.}
\label{tab:allresults}
\setlength{\tabcolsep}{3pt}
\begin{tabular}{@{}l>{\raggedright\arraybackslash}p{0.12\textwidth}>{\raggedright\arraybackslash}p{0.08\textwidth}>{\raggedright\arraybackslash}p{0.15\textwidth}>{\raggedright\arraybackslash}p{0.43\textwidth}>{\raggedright\arraybackslash}p{0.08\textwidth}@{}}
\toprule
ID & Consequence & Status & Setting & Main numbers & Location \\
\midrule
P1 & Blocking in recurrences & [D] Prop.~\ref{prop:block}, \ref{prop:gated_block} & A; elem.\ ($k=4$), dense, and renorm.\ dense ($\alpha=0.3, p_1=p_2=8$, 5 seeds) & Closed form $1-(1-\alpha)^{p_1}=0.9424$. Delta \EncSparseDelta, PH \EncSparsePh, Hebbian, decay, Oja $0.000\pm0.000$. $\alpha$-sweep (5 values): float32 precision (\DeltaAlphaMaxErr). Dense: delta \EncDenseDelta, PH \EncDensePh, Hebbian \EncDenseHebb; softmax $0.000$. Renorm.\ dense: delta \EncDenseNormDelta{} (Eq.~\ref{eq:blocknorm}: $0.6663$), PH \EncDenseNormPh, Hebbian \EncDenseNormHebb. PH over 36 settings: $[0.97, 1.00]$, median $1.00$. Gated delta ($g=0.9$): $0.712$ to $0.156$ for $p_2=1$ to $16$. & \S\ref{sec:experiments}; Fig.~\ref{fig:conditioning}a; App.~\ref{app:proofs}, \ref{app:extra} \\
P2 & Blocking survives training & [M] & T1/T2 elem.\ and dense-trained; in-range summation ($4+4$, probe 12; 10 protocol seeds, $B=1024$) & Cosine, T1: delta \CosDeltaLone, PH \CosPhLone, Hebbian \CosHebbLone, decay \CosDecayLone, Oja \CosOjaLone{} (base cosine $0.82$ delta, $0.85$ Hebbian at 8 pairs). Summation, T1: delta $0.259\pm0.003$, PH $0.278\pm0.002$, Hebbian $0.020\pm0.002$, decay $0.013\pm0.003$, Oja $0.011\pm0.001$, softmax $0.006\pm0.003$; delta$-$Hebbian $+0.240$ $[+0.222, +0.257]$. Delta$-$Hebbian summation across other families: T2 elem.\ $+0.232$ $[+0.031, +0.433]$; T2 dense-trained $+0.209$ $[+0.154, +0.264]$; T1 dense-trained $+0.239$ $[+0.227, +0.251]$. T2 elem.\ cosine: delta \CosDeltaSP{} vs.\ Hebbian \CosHebbSP. & \S\ref{sec:experiments}; Fig.~\ref{fig:conditioning}b; App.~\ref{app:extra} \\
 & \ldots\ phase-1 length & [M] & Position-matched $1+4$, $2+4$, $4+4$, probe 12 & Delta $(4,4)-(1,4)$: $+0.087$ $[+0.082, +0.093]$ (T1 elem.) to $+0.145$ $[+0.140, +0.151]$ (T1 dense-trained); contiguity \& softmax $-0.001$ to $+0.015$. T1 elem.\ ratios $0.664 / 0.899$ vs.\ $0.698 / 0.914$ at mean rate $\alpha=0.68$. & \S\ref{sec:experiments}; App.~\ref{app:extra} \\
 & \ldots\ dense factorial & [M] & $2^3$ factorial, dense cues, 48 models & Cosine suppression tracks $d_{\text{stim}}=32$ (dimension effect $-0.050$ $[-0.095, -0.006]$); control cosine at ceiling ($0.96$--$0.99$); head-level blocking ($0.365$ $[0.296, 0.434]$) and summation ($+0.209$ $[+0.154, +0.264]$) remain intact. & \S\ref{sec:experiments}; App.~\ref{app:extra} \\
 & \ldots\ cue overlap & [M] & T1, test-cue overlap $\rho\in[0,0.6]$ & Head-level $0.47\to 0.28$; cosine $0.084\to 0.025$; summation $0.259\to 0.170$; delta$-$Hebbian CIs exclude $0$ at every $\rho$. & \S\ref{sec:experiments}; App.~\ref{app:extra} \\
P2$'$ & Backward blocking & [D] Prop.~\ref{prop:backward} & A; $n_1=n_2=8$, $d=128$, 5 seeds $\times$ 512 & Ridge/MesaNet ($\rho=0, \lambda=10^{-3}$): $0.49997\to 0.00012$ (normalized $\rho=0.3$: $0.806\to 0.612$). DeltaNet: $0.500\to 0.500$ ($\rho=0$), or rises to $0.641$ ($\rho=0.3$) and $0.783$ ($\rho=0.6$). & \S\ref{sec:predictions}; App.~\ref{app:backward} \\
P3 & Extinction needs error & [D] Prop.~\ref{prop:ext}; [M] & A dense $d=64$ ($10+10$); T1 in-range ($8+8$, probe 16) & A (\% lost): delta $97$, PH $100$, softmax $50$, decay $18$ (passive), Hebbian $0$, Oja $0$. T1 in-range: delta $13.9\pm0.4$ (filler arm $9.3\pm0.7$), PH $12.3\pm0.6$, softmax $94.2\pm0.5$, contiguity $\le0.4$; cue-specific ($10.5\pm0.7$ vs.\ $1.6\pm0.1$). PH over 36 settings: $[99, 100]\%$. & \S\ref{sec:exp_ext}; Fig.~\ref{fig:extinction}a; App.~\ref{app:extra} \\
P4 & Override persistence & [D] Prop.~\ref{prop:repair}; [M] & A elem.; T1 ($5+5$, probe $\le 17$) & A ($p=4 / 8$): Hebbian $1.000/1.000$, decay $0.922/0.851$, Oja $0.991/0.982$, softmax $0.500$, delta $0.240/0.058$, gated delta $0.221/0.049$, PH $0.026/0.0007$. T1 persistence: Hebbian $0.689$, decay $0.668$, Oja $0.676$, softmax $0.676$, PH $0.356$, delta $0.065$; delta$-$Hebbian $-0.624$ $[-0.636, -0.611]$. & \S\ref{sec:exp_ext}; App.~\ref{app:override}, \ref{app:extra} \\
\bottomrule
\end{tabular}
\end{table}

\begin{table}[tp]
\centering\scriptsize
\caption{Consolidated results (continued): recovery, generalization, capacity, and PH-attention.}
\label{tab:allresults_cont}
\setlength{\tabcolsep}{3pt}
\begin{tabular}{@{}l>{\raggedright\arraybackslash}p{0.12\textwidth}>{\raggedright\arraybackslash}p{0.08\textwidth}>{\raggedright\arraybackslash}p{0.15\textwidth}>{\raggedright\arraybackslash}p{0.43\textwidth}>{\raggedright\arraybackslash}p{0.08\textwidth}@{}}
\toprule
ID & Consequence & Status & Setting & Main numbers & Location \\
\midrule
P5 & Recovery \& renewal & [D] Prop.~\ref{prop:ext}; [M] & A dense; T1--T2 (60 models, 20 settings); equivalence \& ABA & A: largest rise \RecoveryMaxRisePct\%. T2 uncontrolled delta: \RecUncontrolledDelta\% (max \RecUncontrolledDeltaMax\%). Controlled T1--T2: max cell rise \RecMaxAny\% (worst model \RecPerModelMax\%, control \RecPerModelMaxNovel\%). Equivalence passes in 20/21 settings (5/6 with $\ge10\%$ extinction). ABA renewal: error-correcting models show no renewal beyond summation in 18/21 models; softmax renews $+38.8\pm0.9\%$ (1L) and $+43.8\pm0.9\%$ (2L). & \S\ref{sec:exp_ext}; Fig.~\ref{fig:extinction}b; App.~\ref{app:extra} \\
P6 & Generalization gradient & [D] functional form & A, $d=64$, seed 0, batch $2{,}048$, 5 conditions & Exponential beats linear in \GenExpWins{} conditions ($R^2=\GenSoftExpR$ vs.\ \GenSoftLinR{} at \GenSoftRange$\times$; saturated $R^2=\GenSatExpR$ vs.\ \GenSatLinR{} at \GenSatRange$\times$). Linear attn.\ $R^2=\GenLinR$. & \S\ref{sec:experiments}; Fig.~\ref{fig:conditioning}c; App.~\ref{app:extra} \\
P7 & Capacity of one state & [D] Thm.~\ref{thm:cap}; [M] & A: sphere keys, $d_k$ $16$--$256$ (and $1024$); \texttt{elu+1} & Per-item (Fig.~\ref{fig:capacity}): Hebbian \CapFidHebb{} (fidelity), \CapIdeHebb{} (id.); delta id.\ \CapIdeDelta, decay id.\ \CapIdeDecay. All-items unrescaled (Thm.~\ref{thm:cap}): $1.08$ $[1.06, 1.08]$ / $1.83$ $[1.82, 1.86]$; ratio $0.38/0.52/0.69$ at $d_k=64/256/1024$. Uncentered \texttt{elu+1}: mean overlap $0.685$, $n^\ast=1$ (fidelity), $2$--$5$ (id.); centering restores sphere exponents. & \S\ref{sec:exp_cap}; Fig.~\ref{fig:capacity}a--b; App.~\ref{app:extra} \\
P7$'$ & Partition across heads & [M] & T1 $H\times d_{\text{head}}$ grid, 3 seeds, 24 pairs & $R^2$ of log error on log state size \GridRtwoStateDelta{} (delta) / \GridRtwoStateHebb{} (Hebb), vs.\ $\log H$ \GridRtwoHeadsDelta/\GridRtwoHeadsHebb. Redundant prediction (Hebb): \HeadErrHoneHebb$\to$\HeadPredExpHebb{} at $H=8$; measured \HeadErrHeightHebb. 4-head diagnostics: ablation $+0.05$--$0.06$, per-item $|r|\le0.03$, output cosine $0.09$--$0.22$. & \S\ref{sec:exp_cap}; Fig.~\ref{fig:capacity}c; App.~\ref{app:extra} \\
P8 & PH volatility tracking & [D] Prop.~\ref{prop:vol}; [M] & A assay; T2 switch \& mixed reliability (noise $0.35$ \& $0.1$) & Rate ratio \VolRatioPhcue{} vs.\ \VolRatioDelta. Switch clean: PH \SwPhClean, delta \SwDeltaClean. Mixed reliability (noise $0.35$, 10 seeds): PH \VolTaskPh, delta \VolTaskDelta, paired \PHPairedDiff{} ($[-0.0099,+0.0025]$); oracle $+0.0718$; $w>0$ $+0.0002$. Lower noise ($0.1$, slow trace): $+0.0033$ $[+0.0025, +0.0042]$ in 10/10 seeds ($14.3\%$ of oracle $+0.0232$), $w=0.161$. & \S\ref{sec:exp_ph}; Fig.~\ref{fig:conditioning}d; App.~\ref{app:extra} \\
\bottomrule
\end{tabular}
\end{table}

\begin{table}[tp]
\centering\scriptsize
\caption{What each entry in Table~\ref{tab:dict} preserves from its classical counterpart and what falls outside the mapping (tiers E, R, I).}
\label{tab:preserved}
\setlength{\tabcolsep}{4pt}
\begin{tabular}{@{}>{\raggedright\arraybackslash}p{0.18\textwidth}>{\raggedright\arraybackslash}p{0.38\textwidth}>{\raggedright\arraybackslash}p{0.40\textwidth}@{}}
\toprule
Entry & Preserved & Outside the correspondence \\
\midrule
Linear attention = Hebbian contiguity (E) & Every co-occurrence adds $x^\top\lambda$; no cue competition and no extinction; retrieval acts as a test trial. & Retrieval may use a query projection distinct from the write key. $\phi=\mathrm{elu}+1$ places trained keys in the positive orthant, so distinct cues are non-orthogonal. \\
DeltaNet = Rescorla--Wagner (E) & Summed-error update, cue competition, asymptote at $\lambda$, signed strengths (conditioned inhibition), extinction as unlearning, independent US dimensions, and salience $\alpha_i=\alpha x_i^2$. & $\beta_t$ is token-computed rather than fixed per CS/US. In trained networks keys are $\phi$-mapped and $\ell_2$-normalized, making single cues equal-salience and compounds non-additive; query and key projections can differ. \\
Decay rule = contiguity + exponential forgetting (E) & Exact identity to exponential strength decay. & Distinct from Wagner's SOP \citep{vogel2019sop,uribe2019quantitative}, where decay acts on stimulus activity states and produces blocking; here blocking is $0$. \\
Gated DeltaNet (R) & Rescorla--Wagner exactly when $\alpha_t\equiv 1$. & When $\alpha_t<1$, global multiplicative forgetting acts before error computation, making blocking shrink with compound length $p_2$ (Prop.~\ref{prop:gated_block}). \\
Titans, linear memory (R) & Rescorla--Wagner exactly when $\eta_t=\alpha_t=0$ (rate $2\theta_t$). & Momentum (signed error history added to write), weight decay on undecayed state, and data-dependent gates have no classical counterpart. \\
Titans, MLP memory (---) & --- & Nonlinear online regression over shared hidden weights breaks elemental summation; RW predictions do not transfer automatically. \\
Oja-type rule (I) & Contiguity plus a $-y^2w$ homeostatic term. & $y$ is clamped to the US rather than the unit's own output, making it an Oja-inspired variant. \\
PH-attention (I, RW--PH hybrid) & Associability tracks recent unsigned prediction error via exponential averaging with one-trial lag, indexed per feature. & Retains the RW error term (unlike PH 1980's $S\alpha\lambda$); applies one shared scalar rate per head and token; normalizes by feature mean; learns slope sign ($w>0$ PH, $w<0$ Mackintosh-like, $w=b=0$ DeltaNet). \\
Softmax (---) & --- & Non-parametric normalized similarity over stored tokens; no finite-state write rule. \\
\bottomrule
\end{tabular}
\end{table}

\FloatBarrier
\section{Proofs and Derivations}
\label{app:proofs}

Throughout, $\Sm\in\R^{d_k\times d_v}$ denotes the attention state, cues are row vectors, and $u\in\R^{d_v}$ is a unit US vector. We write $a,b,c$ for orthonormal elementary cue vectors.

\subsection{Theorem~\ref{thm:rw} (Delta-Rule Attention is Rescorla--Wagner)}
\label{app:thm_rw}
Rescorla--Wagner \citep{rescorla1972theory} updates associative strengths $V$ on a trial with cue vector $x$ and US $\lambda$ via $\Delta V=\alpha\beta\,x^\top(\lambda - xV)$. Substituting $x=\phi(\kv_t)$, $\lambda=\vv_t$, $V=\Sm_{t-1}$, and $\alpha\beta=\beta_t$ yields $\Sm_t=\Sm_{t-1}+\beta_t\phi(\kv_t)^\top(\vv_t-\phi(\kv_t)\Sm_{t-1})$, which is Eq.~\ref{eq:delta}. The converse substitution is immediate. When $\lambda$ is vector-valued, the update decouples across columns of $\Sm$ into independent multi-outcome Rescorla--Wagner updates. $\square$

\subsection{Proposition~\ref{prop:block} (Closed-Form Blocking)}
\label{app:prop_block}
Consider $p_1$ pairings of the phase-1 cue with $u$, then $p_2$ pairings of the compound with $u$, followed by probe $b$ with response $R=\langle b\Sm,u\rangle$.

\paragraph{(i) Hebbian, decay, and Oja rules.}
Under the Hebbian rule, every trial adds $k^\top\otimes u$. In the blocked condition $\Sm=p_1 a^\top\otimes u + p_2(a+b)^\top\otimes u$, so $b\Sm=(p_1\langle b,a\rangle+p_2\langle b,a+b\rangle)u=p_2 u$ since $b\perp a$ and $\|b\|=1$. Replacing $a$ with $c\perp\{a,b\}$ in the control gives $b\Sm=p_2 u$ identically, so $1-R_{\text{blocked}}/R_{\text{control}}=0$. For the decay rule, phase-1 writes remain orthogonal to $b$ and phase-2 writes receive identical decay weights in both conditions, again giving $0$. For Oja's rule, the $-y^2\Sm$ term scales $\Sm$ uniformly and is condition-independent to leading order. $\square$

\paragraph{(ii) Delta rule, additive compound.}
Let $c_+=a+b$ ($\|c_+\|^2=2$) and rate $\alpha\in(0,1)$. For a unit cue $k$, $k\Sm_t=(1-\alpha)k\Sm_{t-1}+\alpha u$, so after $p_1$ trials $a\Sm=(1-\varepsilon)u$ with $\varepsilon:=(1-\alpha)^{p_1}$, while $b\Sm=0$. In phase~2 all updates lie along $c_+^\top$, so we may write the phase-2 contribution as $c_+^\top\otimes w$, giving $R=\langle w,u\rangle$ because $\langle b,c_+\rangle=1$. Since $\|c_+\|^2=2$, $c_+\Sm_t=(1-2\alpha)c_+\Sm_{t-1}+2\alpha u$. In the blocked condition $c_+\Sm_0=(1-\varepsilon)u$, yielding $c_+\Sm_t=u-\varepsilon(1-2\alpha)^t u$; the $t$-th update adds $\alpha\varepsilon(1-2\alpha)^{t-1}c_+^\top\otimes u$, so summing the geometric series gives
\begin{equation}
R_{\text{blocked}}=\alpha\varepsilon\sum_{t=1}^{p_2}(1-2\alpha)^{t-1}=\tfrac{\varepsilon}{2}\bigl(1-(1-2\alpha)^{p_2}\bigr).
\end{equation}
In the control ($c\perp\{a,b\}$), $c_+\Sm_0=0$, giving the same expression with $\varepsilon=1$: $R_{\text{control}}=\tfrac{1}{2}(1-(1-2\alpha)^{p_2})$. Hence
\begin{equation}
\text{blocking index}=1-\frac{R_{\text{blocked}}}{R_{\text{control}}}=1-\varepsilon=1-(1-\alpha)^{p_1},
\label{eq:blockexact}
\end{equation}
independent of $p_2$. $\square$

\paragraph{(iii) Delta rule, renormalized compound.}
If instead $c_\times=(a+b)/\sqrt{2}$ ($\|c_\times\|=1$), write $\beta_0:=c_\times\Sm_0\cdot u=(1-\varepsilon)/\sqrt{2}$ in the blocked condition and $0$ in the control. Then $c_\times\Sm_t=u(1-(1-\beta_0)(1-\alpha)^t)$, and since $\langle b,c_\times\rangle=1/\sqrt{2}$,
\begin{equation}
\text{blocking index}=1-\frac{\bigl(1-(1-\beta_0)(1-\alpha)^{p_2}\bigr)-\beta_0}{1-(1-\alpha)^{p_2}} .
\label{eq:blocknorm}
\end{equation}

\subsection{Proposition~\ref{prop:gated_block} (Gated-Delta Blocking)}
\label{app:gated_block}
For Gated DeltaNet with constant gate $g\in(0,1]$ and rate $\eta$, $\Sm\leftarrow g\Sm+\eta\,k^\top(u-k\,g\Sm)$. With orthonormal $a,b,c$ and compound $a+b$, define $\rho_1=g(1-\eta)$, $\rho_2=g(1-2\eta)$, and $A=\eta(1-\rho_1^{p_1})/(1-\rho_1)$, the response to $A$ after phase~1. During phase~2, $m_t=\langle(a+b)\Sm_t,u\rangle$ and $r_t=\langle b\Sm_t,u\rangle$ evolve as
\begin{equation}
m_t=\rho_2 m_{t-1}+2\eta,\qquad r_t=g\,r_{t-1}+\eta(1-g\,m_{t-1}),
\end{equation}
with $r_0=0$, $m_0=A$ (blocked), and $m_0=0$ (control). By linearity in the initial state difference $m_0$,
\begin{equation}
r^{\mathrm{ctl}}_{p_2}-r^{\mathrm{blk}}_{p_2}=\eta\,g\,A\sum_{t=1}^{p_2}g^{p_2-t}\rho_2^{t-1},
\end{equation}
which yields Eq.~\ref{eq:blockgated_main}. At $g=1$ this reduces to Eq.~\ref{eq:blockexact}; for $g<1$ it is positive and shrinks as $p_2$ grows (Table~\ref{tab:gated}). Evaluating a gated recurrence copy (elementary-feature cues, $d=128$, $\eta=0.3$, $p_1=8$, 5 seeds $\times$ 512) matches Eq.~\ref{eq:blockgated_main} within $2.0\times 10^{-7}$ in every cell. $\square$

\begin{table}[t]
\centering\small
\caption{Blocking index of the gated delta rule (Eq.~\ref{eq:blockgated_main}; $\eta=0.3$, $p_1=8$).}
\label{tab:gated}
\begin{tabular}{@{}lcccc@{}}
\toprule
$g$ & $p_2=1$ & $p_2=4$ & $p_2=8$ & $p_2=16$ \\
\midrule
1.0 (Prop.~\ref{prop:block}) & 0.942 & 0.942 & 0.942 & 0.942 \\
0.98 & 0.890 & 0.848 & 0.783 & 0.666 \\
0.90 & 0.712 & 0.548 & 0.363 & 0.156 \\
\bottomrule
\end{tabular}
\end{table}

\paragraph{Token-dependent decay gates.} If a decay gate depends only on the current token (scalar per head or per channel), blocking under Proposition~\ref{prop:block}'s assumptions remains identically $0$: phase-1 writes never touch $b$'s rows, and phase-2 tokens—and thus phase-2 gates—are identical across blocked and control conditions (verified numerically with $\sigma(w\cdot[k_t;v_t]+1)$, gate s.d.\ $0.36$, returning $0.000$ for both gate types).

\paragraph{Numerical verification of Proposition~\ref{prop:block}.} Sweeping $\alpha\in\{0.1,0.2,0.3,0.4,0.5\}$ ($p_1=p_2=8$) matches Eq.~\ref{eq:blockexact} to float32 rounding (maximum error \DeltaAlphaMaxErr). Under elementary-feature cues, Hebbian (\EncSparseHebb), decay (\EncSparseDecay), and Oja (\EncSparseOja) sit at zero. With dense random cues ($d=128$, typical overlap $\approx 0.09$) and a renormalized compound, Eq.~\ref{eq:blocknorm} predicts $0.6663$ and simulation measures \EncDenseNormDelta.

\subsection{Proposition~\ref{prop:backward} (Backward Blocking)}
\label{app:backward}
\emph{Proof.} (i) Phase-2 writes lie along $a^\top$; when $\langle a,b\rangle=0$, the row read by $b$ is unchanged under Hebbian, DeltaNet, and PH updates. With overlap $\rho>0$, each delta trial changes $\langle b\Sm,u\rangle$ by $\beta_t\rho(1-\langle a\Sm,u\rangle)>0$ while $a$ under-predicts $u$, and Hebbian adds $\rho$ per trial.
(ii) In the basis $(a,b)$, the ridge normal equations after $n_1$ compound trials $(a+b)\to u$ and $n_2$ solo trials $a\to u$ are $\bigl[\begin{smallmatrix}n_1+n_2+\lambda & n_1\\ n_1 & n_1+\lambda\end{smallmatrix}\bigr]V=(n_1+n_2,\,n_1)^\top$, whose second coordinate gives $\lambda n_1/(\lambda^2+\lambda(2n_1+n_2)+n_1n_2)$; after phase~1 alone, symmetry gives $V_a=V_b=n_1/(2n_1+\lambda)$. For the unit-normalized compound $c=(a+b)/s$ ($s=\sqrt{2+2\rho}<2$), as $\lambda\to 0$ the minimum-norm fit after phase~1 is $V=c^\top$ with $\langle b,c\rangle=(1+\rho)/s=s/2$; after phase~2, exact interpolation requires $cV=aV=1$, so $b=sc-a$ gives $bV=s-1<s/2$. $\square$

In a float64 ridge recurrence (initial inverse covariance $I/\lambda$, matched to a direct ridge solve within $1.3\times 10^{-14}$; $n_1=n_2=8$, $d=128$, 5 seeds $\times$ 512), exact least squares reduces the $B$ response across every condition (e.g., additive $\rho=0,\lambda=10^{-3}$: $0.49997\to 0.00012$; normalized $\rho=0.3$: $0.806\to 0.612$), whereas DeltaNet, Hebbian, and PH never reduce it (delta additive: $0.500\to 0.500$ at $\rho=0$, rising to $0.641$ at $\rho=0.3$ and $0.783$ at $\rho=0.6$).

\subsection{Proposition~\ref{prop:ext} (Extinction and Single-State Non-Recovery)}
\label{app:prop_ext}
Under the Hebbian rule, $(\kv,\mathbf{0})$ adds $\mathbf{0}$, leaving $\Sm$ unchanged. Under decay, $\Sm$ is multiplied by $\gamma$ per step regardless of the presented cue, yielding passive forgetting $\gamma^p$. Under the delta rule, $\alpha\phi(\kv)^\top(\mathbf{0}-\phi(\kv)\Sm)$ drives $\phi(\kv)\Sm\to 0$ geometrically at rate $(1-\alpha\|\phi(\kv)\|^2)$, producing cue-specific extinction.

For recovery, let $R_T=\langle\phi(A)\Sm_T,U\rangle\ge 0$ and suppose subsequent writes $(\kv_s,\vv_s)_{s>T}$ satisfy $\langle\phi(A),\phi(\kv_s)\rangle=0$. Under Hebbian and delta updates, orthogonality leaves $\phi(A)\Sm_t=\phi(A)\Sm_{t-1}$ and hence $R_t=R_{t-1}$; under decay, $R_t=\gamma R_{t-1}$ is non-increasing. Thus no recovery can occur within the orthogonal single-state suffix. $\square$

\subsection{Proposition~\ref{prop:repair} (Override Persistence Under Re-Pairing)}
\label{app:override}
\emph{Proof.} Hebbian: re-pairing $A\to W$ ($W\perp U$) adds $A^\top W$, leaving the $U$-component of $A\Sm$ untouched ($1$). Decay: every step multiplies $\Sm$ by $\gamma$, yielding $\gamma^{p_2+n}$. Delta: for unit cue $A$, $A\Sm_t=(1-\beta)A\Sm_{t-1}+\beta W$, attenuating the $U$-component by $(1-\beta)^{p_2}$, while orthogonal intervening keys leave $A\Sm$ unchanged. Gated delta: $A\Sm_t=(1-\beta)\gamma A\Sm_{t-1}+\beta W$ during re-pairing and multiplies by $\gamma$ on subsequent orthogonal steps, giving $((1-\beta)\gamma)^{p_2}\gamma^n$. PH matches delta with effective rate $\bar b$. Softmax averages equally over all identical $A$ tokens ($p_1$ copies of $U$ and $p_2$ of $W$), giving $p_1/(p_1+p_2)$. $\square$

All closed forms match simulated recurrences to float32 precision across 60 of 60 $(p_1,p_2)$ and encoding cells (maximum deviation $4.2\times 10^{-7}$ for Hebbian, decay, delta, and softmax; $5.9\times 10^{-7}$ for gated delta).

\subsection{Theorem~\ref{thm:cap} (Two Capacity Regimes)}
\label{app:thm_cap}
Querying $\Sm=\sum_{j=1}^n \kv_j^\top\vv_j$ (independent unit-spherical $\kv_j\in\R^{d_k}$, $\vv_j\in\R^{d_v}$) with $\kv_m$ gives
\begin{equation}
\mathbf{r}_m=\kv_m\Sm=\underbrace{\vv_m}_{\text{signal}}+\underbrace{\sum_{j\ne m}c_j\vv_j}_{\text{noise}},\qquad c_j=\langle\kv_m,\kv_j\rangle.
\end{equation}

\paragraph{(1) Fidelity.} Let $N_m:=\sum_{j\ne m}c_j\vv_j$ and $A_m:=\sum_{j\ne m}c_j^2$. Because $\E[c_j]=0$, $\E[c_j^2]=1/d_k$, and $\vv_j$ are independent of $\kv_j$, $\E\|N_m\|^2=(n-1)/d_k$ exactly. Since $\sqrt{d_k}\langle z,g\rangle$ is sub-Gaussian for spherical $g$, $d_k c_j^2$ is sub-exponential, so Bernstein's inequality \citep{vershynin2018high} gives for $t\ge 1$:
\begin{equation}
A_m\le\frac{n-1}{d_k}\left[1+C\left(\sqrt{\frac{t}{n}}+\frac{t}{n}\right)\right]
\label{eq:keyconc}
\end{equation}
with probability at least $1-2e^{-t}$. Conditioned on the keys, applying the dependent-coordinate Hanson--Wright inequality \citep{hanson1971bound, adamczak2015note} to $\|N_m\|^2$ yields
\begin{equation}
\|N_m\|^2\le A_m\left[1+C\left(\sqrt{\frac{t}{d_v}}+\frac{t}{d_v}\right)\right]
\label{eq:valueconc}
\end{equation}
with probability at least $1-2e^{-t}$. Setting $t=\log(4n^2/\epsilon)$ and taking a union bound over all $m\in\{1,\dots,n\}$ yields Eq.~\ref{eq:fidcond}. Lower-tail bounds confirm that when $d_v\gtrsim\log d_k$, loads above a constant multiple of $\delta d_k$ fail even for a single item with non-vanishing probability, giving $n^\ast=\Theta(\delta d_k)$. $\square$

\paragraph{(2) Identification.} Correctly selecting $\vv_m$ over $i\ne m$ is equivalent to positivity of the pairwise margin
\begin{equation}
D_{mi}:=\langle\mathbf{r}_m,\vv_m-\vv_i\rangle = 1-\langle\vv_m,\vv_i\rangle + c_i\bigl(\langle\vv_i,\vv_m\rangle-1\bigr) + \sum_{j\ne i,m}c_j\langle\vv_j,\vv_m-\vv_i\rangle.
\end{equation}
Conditioned on $(\kv_m,\vv_m,\vv_i)$, terms in the sum are independent products of centered sub-Gaussian variables and hence sub-exponential. Bernstein's inequality combined with the sub-Gaussian tails of the first two random terms yields constants $c,C>0$ such that
\begin{equation}
\Pr(D_{mi}\le 0)\le 2\exp\!\left[-\frac{c}{d_k^{-1}+d_v^{-1}+n(d_k d_v)^{-1}}\right].
\end{equation}
Taking a union bound over $<n^2$ ordered pairs $(m,i)$ yields Eq.~\ref{eq:idcond}; for $d_v=d_k$ and $n\gg d_k$, $n\log(n/\epsilon)\lesssim d_k^2$ suffices, establishing $n^\ast=\Omega(d_k^2/\log(d_k/\epsilon))$. $\square$

\subsection{Proposition~\ref{prop:vol} (Volatility Tracking)}
\label{app:prop_vol}
Under matched token-computed base gates $\sigma(\xv_t\W_\beta)$, the delta rule uses identical rates for both cues, and the global-associability control uses a single shared scalar trace, giving a rate ratio of $1$. For PH-attention, common factors cancel in the ratio:
\begin{equation}
\frac{\beta^{\mathrm{vol}}}{\beta^{\mathrm{sta}}} = \frac{\sigma(w\alpha^{\mathrm{vol}}+b)}{\sigma(w\alpha^{\mathrm{sta}}+b)} > 1
\end{equation}
whenever $\alpha^{\mathrm{vol}}>\alpha^{\mathrm{sta}}$ and $w>0$, by strict monotonicity of $\sigma$. $\square$

\subsection{Multi-Head Error Bounds}
\label{app:err_remark}
\begin{remark}[Error propagation under redundant vs.\ specialized heads]
\label{thm:err}
For an $L$-layer network with $H$ heads per layer and per-head failure probability $p$: (i) if heads are redundant and fail independently, $r\lesssim Lp^H$; (ii) if heads are specialized and all required downstream, $r\lesssim LHp$.
\end{remark}
Both bounds follow directly from union bounds: in (i) a layer fails only if all $H$ heads fail ($p^H$), whereas in (ii) a layer fails if any head fails ($\le Hp$). Because they differ exponentially in $H$ and (i) relies on independent failures despite shared residual inputs, Section~\ref{sec:exp_cap} evaluates the relationship empirically.

\FloatBarrier
\section{Experimental Details}
\label{app:exp}

\paragraph{Attention variants.} All linear-attention variants use readout $\mathbf{r}_t=\phi(\qv_t)\Sm_t$ with $\phi=\mathrm{elu}(\cdot)+1$ and differ only in the write rule (Table~\ref{tab:dict}). Trained Oja models use fixed rate $0.5$. In trained DeltaNet, PH-attention, and global-associability models, keys are $\ell_2$-normalized to ensure contractive updates \citep{yang2024parallelizing}. The associability denominator in Eq.~\ref{eq:ph_alpha} uses a numerical floor of $10^{-6}$ in trained models and standard recurrences ($10^{-9}$ in the analytic volatility sweep). In analytic blocking recurrences, keys are unnormalized so that additive compounds satisfy $\|a+b\|^2=2$ and $\alpha\|k\|^2<2$. Decay models spread per-head $\gamma$ geometrically over $[0.6, 0.999]$ as in RetNet. Softmax uses standard causal dot-product attention.

\paragraph{Analytic experiments.} In the encoding-robustness and PH-sensitivity suites, elementary-feature cues have disjoint random supports of size $k=4$, and dense cues sample unit-norm random vectors in $d=64$ (or $d=128$ where stated). Conditioning parameters are $\alpha=\beta=0.3$, $\gamma=0.98$, $p_1=p_2=8$ unless stated (extinction uses $10+10$ trials; overshadowing uses 12 compound trials with salience ratio $r=3$ and $\alpha=\beta=1/(r^2+1)$); default PH constants are $w=4, b=-1, \eta=0.7$ (and $w=0.75, b=-1.5, \eta=0.5$ in the volatility assay, averaged over 2,048 examples and 5 seeds across 12 alternating blocks). Batch size is 512 (5 seeds for blocking; 3 for extinction and overshadowing; seed 0, batch 2,048 for generalization sweeps and SNR).

\paragraph{Capacity procedure.} Memories store $n$ i.i.d.\ spherical pairs $(\kv_j,\vv_j)$ ($d_v=d_k$, no feature map) across 5 seeds $\times$ 64 memories (4,000 memories per $d_k$ for theorem-matched all-items fidelity; 1,000 for all-items identification), using doubling and integer bisection to find the largest $n^\ast$ at 95\% success. Under the per-item criterion (Figure~\ref{fig:capacity}), each retrieved vector is rescaled by its per-item least-squares scalar ($\delta=0.25$, corresponding to $\cos>0.866$ and $\delta'=\delta/(1-\delta)=1/3$). Exponents are OLS slopes of $\log n^\ast$ on $\log d_k$ over $d_k\in[16,256]$; bootstrap CIs use 2,000 memory replicates.

\paragraph{Trained models.} Inputs are continuous tokens $[\text{cue};\text{outcome};\text{probe-flag}]$ with outcome zeroed at probes. Models use $d_{\text{model}}=128$, 4 heads, 1 or 2 layers, no MLP (isolating associative memory), pre-LayerNorm, and a linear head, trained with cosine loss at probe positions on random associative sets of 4--16 pairs (AdamW, weight decay $0.01$, grad clip $1.0$, one-cycle peak LR $3\times 10^{-4}$, batch size 128, $6{,}000$--$10{,}000$ steps, 3 model seeds per condition). Conditioning protocols never appear during training. Because training sequences have length $\le 20$, positional embeddings beyond index 19 stay at initialization; all trained conditioning probes are placed at index $\le 19$ (cosine blocking at 16, extinction at 16, overshadowing at 12, generalization at 8, summation at 12, override persistence at $\le 17$). For phase-1 trial-count sweeps, $(1,4)$ and $(2,4)$ designs are padded with leading disjoint filler pairs so phase~2 and the probe sit at fixed positions (index 12).

\paragraph{Trained blocking readouts.} We report both (i) the scale-invariant cosine blocking index and (ii) the magnitude-free $B{+}G$ summation test (Section~\ref{sec:experiments}, 10 protocol seeds, batch 1,024).

\paragraph{$k$-hop, $(H,d_{\mathrm{head}})$ grid, and distribution-shift tasks.}
The $k$-hop task samples 10 distinct symbols from a 48-symbol vocabulary forming a single-cycle successor map (eliminating distractor ambiguity), trained for 20k steps at LR $10^{-3}$ with MLP blocks (3 seeds). The $(H,d_{\mathrm{head}})$ grid sweeps $H\in\{1,2,4,8\}\times d_{\text{head}}\in\{8,16,32,64\}$ (1 layer, 24 stored pairs, 3 seeds; plus a 6-pair below-capacity sweep). Regime switch re-pairs all cues mid-context (noisy variant corrupts 20\% of outcomes). Mixed reliability interleaves volatile cues (fresh outcome each trial) with stable cues (fixed clean outcome plus additive Gaussian noise of std $0.35$, or $0.10$ in the lower-noise follow-up; 10 seeds).

\FloatBarrier
\section{Additional Results}
\label{app:extra}

\paragraph{SNR and encoding robustness.} Measured SNR matches $d_k/(n-1)$ to \SnrRelErrPct\% mean relative error across $d_k\in\{32,64,128,256\}$ and $n\in[8,128]$. Blocking holds across elementary-feature (delta \EncSparseDelta, PH \EncSparsePh, Hebbian \EncSparseHebb, decay \EncSparseDecay, Oja \EncSparseOja), dense (delta \EncDenseDelta, PH \EncDensePh, Hebbian \EncDenseHebb), and renormalized dense compounds (delta \EncDenseNormDelta, PH \EncDenseNormPh, Hebbian \EncDenseNormHebb).

\paragraph{Trained key geometry and mechanistic gap ladder.} Instrumenting a verbatim copy of the trained one-layer mixer (Table~\ref{tab:geometry}) shows that disjoint elementary inputs reach key cosine $0.219$ under DeltaNet ($0.479$ centered on null key $k_0$) and partial compound additivity ($0.852$). Mean delta write rates are $0.678\pm 0.053$ in phase~1 ($0.69$--$0.75$ across 11 of 12 heads; one head inactive at $0.076$) and $0.502\pm 0.041$ on compounds; PH gates $2\sigma(w\alpha_t+b)$ damp protocol writes ($0.61$ in phase~1, $0.71$ on compounds, $1.00$ at probe). Table~\ref{tab:ladder} in Section~\ref{sec:experiments} decomposes the gap from Proposition~\ref{prop:block}'s closed form ($0.956$ at per-head rates) to exact head replay ($0.466$) and final cosine output ($0.083$).

\begin{table}[t]
\centering\scriptsize
\caption{Trained key geometry in one-layer elementary-feature models (mean over 3 models). $k_0$: null-cue key.}
\label{tab:geometry}
\setlength{\tabcolsep}{3pt}
\begin{tabular}{@{}lccccc@{}}
\toprule
Quantity & delta & PH & Hebb & decay & Oja \\
\midrule
$\cos(k_A,k_C)$ & $0.219$ & $0.208$ & $0.080$ & $0.295$ & $0.093$ \\
centered $\cos(k_A{-}k_0,k_C{-}k_0)$ & $0.479$ & $0.477$ & $0.079$ & $0.193$ & $0.091$ \\
$\cos(k_{A+B},\mathrm{norm}(k_A{+}k_B))$ & $0.852$ & $0.847$ & $0.867$ & $0.903$ & $0.874$ \\
$\|\phi(k_{A+B})\|/\|\phi(k_A)\|$ & $1.543$ & $1.576$ & $1.103$ & $1.083$ & $1.145$ \\
$\cos(\phi(q_B),k_A)$, phase 1 & $0.175$ & $0.167$ & $0.082$ & $0.287$ & $0.098$ \\
$\cos(\phi(q_B),k_C)$, phase 1 & $0.178$ & $0.169$ & $0.079$ & $0.282$ & $0.092$ \\
\bottomrule
\end{tabular}
\end{table}

\begin{figure}[t]
\centering
\includegraphics[width=0.92\textwidth]{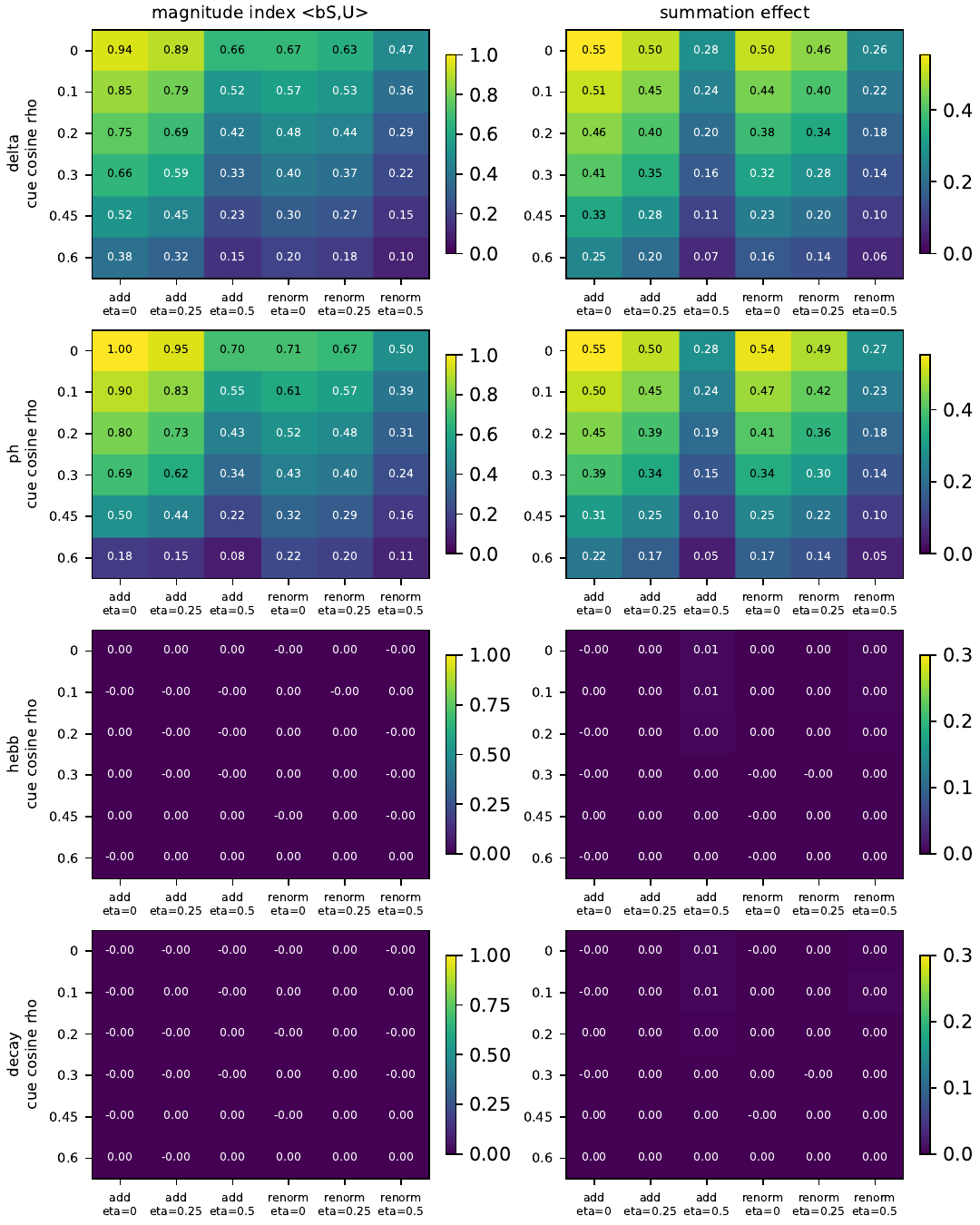}
\caption{Analytic representation sweep across cue overlap $\rho\in[0,0.6]$ and compound geometry (additive vs.\ renormalized, with configural weight $\eta\in\{0,0.25,0.5\}$). DeltaNet and PH remain strictly positive in every cell, while Hebbian and decay stay at $0.00$.}
\label{fig:sweep}
\end{figure}

\paragraph{Departures from orthogonality and additivity.} Figure~\ref{fig:sweep} sweeps cue overlap $\rho\in[0,0.6]$ and configural compound weight $\eta\in[0,0.5]$ in the recurrences ($d_k=128$, 5 seeds $\times$ 1,024); error-correcting blocking remains positive in all 36 cells, while contiguity stays at zero. Tables~\ref{tab:msweep} and \ref{tab:overlap} report the trained elementary-to-dense interpolation sweep (48 newly trained 1-layer models at $d_{\text{stim}}=128$) and controlled test-cue overlap $\rho\in[0,0.6]$ (raising trained key cosine up to $0.550$). Table~\ref{tab:factorial} reports the $2^3$ dense-cue factorial (48 models) confirming that $d_{\text{stim}}=32$ ceiling saturation ($r=-0.90$) accounts for low cosine indices while head-level and summation blocking remain clear. Table~\ref{tab:sumdesigns} compiles summation blocking across all four trained families and trial designs $(p_1,p_2)$.

\begin{table}[tp]
\centering\scriptsize
\caption{Elementary-to-dense cue interpolation at $d_{\text{stim}}=128$ in one-layer models ($n=3$ per rule).}
\label{tab:msweep}
\setlength{\tabcolsep}{3pt}
\resizebox{\textwidth}{!}{%
\begin{tabular}{@{}cccccccc@{}}
\toprule
$m$ & input $\langle|\cos(A,B)|\rangle$ & delta cosine & Hebb cosine & $\Delta$ cosine [95\% CI] & delta summation & Hebb summation & $\Delta$ summation [95\% CI] \\
\midrule
0 & 0.000 & $0.084\pm0.002$ & $0.002\pm0.000$ & 0.083 [0.075, 0.091] & $0.259\pm0.003$ & $0.020\pm0.002$ & $+0.240$ [$+0.222$, $+0.257$] \\
0.25 & 0.031 & $0.082\pm0.001$ & $0.004\pm0.001$ & 0.078 [0.073, 0.083] & $0.255\pm0.003$ & $0.020\pm0.001$ & $+0.235$ [$+0.223$, $+0.247$] \\
0.50 & 0.062 & $0.077\pm0.003$ & $0.007\pm0.000$ & 0.070 [0.059, 0.081] & $0.241\pm0.002$ & $0.018\pm0.003$ & $+0.222$ [$+0.205$, $+0.240$] \\
0.75 & 0.070 & $0.078\pm0.004$ & $0.008\pm0.000$ & 0.070 [0.052, 0.088] & $0.233\pm0.006$ & $0.018\pm0.003$ & $+0.215$ [$+0.198$, $+0.232$] \\
1 (dense) & 0.070 & $0.078\pm0.003$ & $0.008\pm0.001$ & 0.070 [0.058, 0.081] & $0.236\pm0.006$ & $0.022\pm0.002$ & $+0.215$ [$+0.198$, $+0.232$] \\
\bottomrule
\end{tabular}}
\end{table}

\begin{table}[tp]
\centering\scriptsize
\caption{Controlled test-cue overlap $\rho$ in one-layer elementary-feature models ($n=3$ per rule).}
\label{tab:overlap}
\setlength{\tabcolsep}{3pt}
\resizebox{\textwidth}{!}{%
\begin{tabular}{@{}ccccccccc@{}}
\toprule
$\rho$ & $\cos(k_A,k_C)$ & delta cos. & Hebb cos. & $\Delta$ cos.\ [95\% CI] & head-level & delta sum. & Hebb sum. & $\Delta$ sum.\ [95\% CI] \\
\midrule
0.0 & 0.219 & $0.084\pm0.003$ & $0.002\pm0.000$ & 0.082 [0.069, 0.096] & 0.466 (0.515) & $0.259\pm0.003$ & $0.020\pm0.002$ & $+0.240$ [$+0.222$, $+0.257$] \\
0.1 & 0.256 & $0.077\pm0.003$ & $0.001\pm0.001$ & 0.075 [0.061, 0.089] & 0.462 & $0.263\pm0.003$ & $0.021\pm0.003$ & $+0.243$ [$+0.217$, $+0.269$] \\
0.2 & 0.299 & $0.068\pm0.003$ & $0.002\pm0.001$ & 0.066 [0.052, 0.079] & 0.442 & $0.258\pm0.004$ & $0.018\pm0.003$ & $+0.240$ [$+0.213$, $+0.266$] \\
0.3 & 0.349 & $0.056\pm0.003$ & $0.002\pm0.000$ & 0.054 [0.041, 0.068] & 0.412 & $0.245\pm0.005$ & $0.018\pm0.002$ & $+0.227$ [$+0.199$, $+0.255$] \\
0.45 & 0.440 & $0.040\pm0.002$ & $0.001\pm0.000$ & 0.038 [0.029, 0.048] & 0.353 & $0.213\pm0.005$ & $0.016\pm0.001$ & $+0.197$ [$+0.172$, $+0.222$] \\
0.6 & 0.550 & $0.025\pm0.001$ & $0.001\pm0.000$ & 0.025 [0.019, 0.030] & 0.277 & $0.170\pm0.005$ & $0.012\pm0.000$ & $+0.158$ [$+0.139$, $+0.177$] \\
\bottomrule
\end{tabular}}
\end{table}

\begin{table}[tp]
\centering\scriptsize
\caption{Dense-cue $2^3$ factorial ($n=3$ per cell and rule; $\Delta=$ delta$-$Hebbian paired by seed). Lower block isolates $d_{\text{stim}}=32$ ceiling diagnostics.}
\label{tab:factorial}
\setlength{\tabcolsep}{3pt}
\resizebox{\textwidth}{!}{%
\begin{tabular}{@{}lccccc@{}}
\toprule
Cell ($L$, $d_{\text{stim}}$, $p_{\text{comp}}$) & delta cosine & Hebb cosine & $\Delta$ cosine [95\% CI] & $\Delta$ summation, elem.\ [95\% CI] & $\Delta$ summation, dense [95\% CI] \\
\midrule
1, 32, 0 & $0.007\pm0.005$ & $0.000\pm0.000$ & 0.007 [$-0.013$, 0.027] & $+0.242$ [$+0.221$, $+0.263$] & $+0.227$ [$+0.225$, $+0.230$] \\
1, 32, 0.3 & $0.006\pm0.003$ & $0.000\pm0.000$ & 0.005 [$-0.006$, 0.016] & $+0.239$ [$+0.227$, $+0.251$] & $+0.229$ [$+0.222$, $+0.236$] \\
2, 32, 0 & $0.014\pm0.005$ & $0.001\pm0.001$ & 0.013 [$-0.011$, 0.037] & $+0.199$ [$+0.047$, $+0.351$] & $+0.192$ [$+0.048$, $+0.336$] \\
2, 32, 0.3 (baseline) & $0.010\pm0.001$ & $0.001\pm0.000$ & 0.009 [0.005, 0.013] & $+0.209$ [$+0.154$, $+0.264$] & $+0.196$ [$+0.141$, $+0.252$] \\
1, 128, 0 & $0.075\pm0.007$ & $0.005\pm0.001$ & 0.069 [0.032, 0.106] & $+0.207$ [$+0.190$, $+0.225$] & $+0.203$ [$+0.182$, $+0.224$] \\
1, 128, 0.3 & $0.048\pm0.004$ & $0.006\pm0.001$ & 0.043 [0.021, 0.064] & $+0.199$ [$+0.182$, $+0.216$] & $+0.195$ [$+0.194$, $+0.197$] \\
2, 128, 0 & $0.087\pm0.045$ & $0.007\pm0.003$ & 0.080 [$-0.118$, 0.278] & $+0.218$ [$+0.020$, $+0.416$] & $+0.226$ [$+0.064$, $+0.387$] \\
2, 128, 0.3 & $0.029\pm0.000$ & $0.007\pm0.004$ & 0.022 [0.006, 0.037] & $+0.243$ [$+0.235$, $+0.250$] & $+0.248$ [$+0.223$, $+0.274$] \\
\bottomrule
\end{tabular}}

\medskip
\begin{tabular}{@{}lcccc@{}}
\toprule
Cell ($d_{\text{stim}}=32$) & delta head-level, last layer & $\Delta$ magnitude index & $\Delta$ summation, elem. & control cosine \\
\midrule
1, 32, 0 & 0.258 [0.152, 0.363] & 0.329 [0.149, 0.508] & $+0.242$ [$+0.221$, $+0.263$] & 0.980 \\
1, 32, 0.3 & 0.278 [0.195, 0.360] & 0.270 [0.102, 0.437] & $+0.239$ [$+0.227$, $+0.251$] & 0.992 \\
2, 32, 0 & 0.300 [0.027, 0.572] & 0.243 [$-0.092$, 0.577] & $+0.199$ [$+0.047$, $+0.351$] & 0.956 \\
2, 32, 0.3 (baseline) & 0.365 [0.296, 0.434] & 0.151 [0.020, 0.281] & $+0.209$ [$+0.154$, $+0.264$] & 0.991 \\
\bottomrule
\end{tabular}
\end{table}

\begin{table}[tp]
\centering\scriptsize
\caption{In-range summation blocking across all trained model families and $(p_1,p_2)$ designs (mean $\pm$ s.e.m., $n=3$ models; ``+3f''/``+2f'' denote leading filler pairs fixing phase~2 and probe positions).}
\label{tab:sumdesigns}
\setlength{\tabcolsep}{2.2pt}
\resizebox{\textwidth}{!}{%
\begin{tabular}{@{}llcccccccll@{}}
\toprule
Family & Design & Probe & delta & PH & Hebb & decay & Oja & softmax & delta$-$Hebb [95\% CI] & PH$-$Hebb [95\% CI] \\
\midrule
1L elem. & (1,4) & 9 & $0.253\pm0.004$ & $0.318\pm0.003$ & $0.009\pm0.001$ & $0.001\pm0.003$ & $-0.000\pm0.001$ & $0.001\pm0.003$ & $+0.244$ [$+0.232$, $+0.256$] & $+0.308$ [$+0.300$, $+0.317$] \\
 & (2,4) & 10 & $0.263\pm0.003$ & $0.293\pm0.002$ & $0.013\pm0.002$ & $0.006\pm0.003$ & $0.004\pm0.000$ & $0.003\pm0.003$ & $+0.250$ [$+0.238$, $+0.262$] & $+0.280$ [$+0.272$, $+0.288$] \\
 & (1,4)+3f & 12 & $0.172\pm0.003$ & $0.207\pm0.002$ & $0.009\pm0.001$ & $0.001\pm0.003$ & $-0.000\pm0.001$ & $0.001\pm0.003$ & $+0.163$ [$+0.154$, $+0.172$] & $+0.198$ [$+0.189$, $+0.207$] \\
 & (2,4)+2f & 12 & $0.233\pm0.002$ & $0.242\pm0.002$ & $0.013\pm0.002$ & $0.006\pm0.003$ & $0.004\pm0.000$ & $0.003\pm0.003$ & $+0.220$ [$+0.210$, $+0.230$] & $+0.229$ [$+0.216$, $+0.242$] \\
 & (4,4) & 12 & $0.259\pm0.003$ & $0.278\pm0.002$ & $0.020\pm0.002$ & $0.013\pm0.003$ & $0.011\pm0.001$ & $0.006\pm0.003$ & $+0.240$ [$+0.222$, $+0.257$] & $+0.258$ [$+0.243$, $+0.274$] \\
 & (6,4) & 14 & $0.254\pm0.003$ & $0.268\pm0.002$ & $0.024\pm0.003$ & $0.019\pm0.003$ & $0.016\pm0.002$ & $0.009\pm0.003$ & $+0.230$ [$+0.209$, $+0.250$] & $+0.244$ [$+0.226$, $+0.263$] \\
 & (8,4) & 16 & $0.250\pm0.003$ & $0.260\pm0.002$ & $0.028\pm0.003$ & $0.022\pm0.002$ & $0.020\pm0.002$ & $0.011\pm0.003$ & $+0.222$ [$+0.198$, $+0.246$] & $+0.232$ [$+0.211$, $+0.254$] \\
 & (4,7) & 18 & $0.257\pm0.003$ & $0.278\pm0.002$ & $0.015\pm0.002$ & $0.008\pm0.003$ & $0.006\pm0.000$ & $0.004\pm0.003$ & $+0.241$ [$+0.226$, $+0.257$] & $+0.262$ [$+0.249$, $+0.276$] \\
\midrule
2L elem. & (1,4) & 9 & $0.228\pm0.047$ & $0.252\pm0.058$ & $0.006\pm0.002$ & $0.001\pm0.001$ & $-0.004\pm0.002$ & $0.001\pm0.004$ & $+0.222$ [$+0.018$, $+0.426$] & $+0.246$ [$-0.008$, $+0.501$] \\
 & (2,4) & 10 & $0.252\pm0.051$ & $0.253\pm0.060$ & $0.012\pm0.002$ & $0.004\pm0.001$ & $-0.003\pm0.002$ & $0.004\pm0.004$ & $+0.240$ [$+0.020$, $+0.460$] & $+0.241$ [$-0.020$, $+0.502$] \\
 & (1,4)+3f & 12 & $0.152\pm0.031$ & $0.169\pm0.042$ & $0.006\pm0.002$ & $0.001\pm0.001$ & $-0.000\pm0.003$ & $0.002\pm0.004$ & $+0.147$ [$+0.009$, $+0.285$] & $+0.164$ [$-0.019$, $+0.346$] \\
 & (2,4)+2f & 12 & $0.205\pm0.040$ & $0.204\pm0.049$ & $0.012\pm0.002$ & $0.004\pm0.001$ & $0.001\pm0.003$ & $0.003\pm0.004$ & $+0.193$ [$+0.019$, $+0.367$] & $+0.192$ [$-0.022$, $+0.406$] \\
 & (4,4) & 12 & $0.253\pm0.046$ & $0.253\pm0.059$ & $0.021\pm0.002$ & $0.009\pm0.001$ & $0.002\pm0.002$ & $0.007\pm0.004$ & $+0.232$ [$+0.031$, $+0.433$] & $+0.232$ [$-0.021$, $+0.485$] \\
 & (6,4) & 14 & $0.247\pm0.042$ & $0.248\pm0.054$ & $0.028\pm0.002$ & $0.012\pm0.002$ & $0.007\pm0.002$ & $0.009\pm0.004$ & $+0.220$ [$+0.037$, $+0.403$] & $+0.221$ [$-0.012$, $+0.454$] \\
 & (8,4) & 16 & $0.237\pm0.037$ & $0.239\pm0.048$ & $0.032\pm0.002$ & $0.013\pm0.002$ & $0.012\pm0.003$ & $0.012\pm0.004$ & $+0.205$ [$+0.044$, $+0.366$] & $+0.207$ [$+0.001$, $+0.413$] \\
 & (4,7) & 18 & $0.250\pm0.042$ & $0.255\pm0.058$ & $0.013\pm0.002$ & $0.000\pm0.001$ & $-0.006\pm0.003$ & $0.005\pm0.004$ & $+0.237$ [$+0.052$, $+0.422$] & $+0.242$ [$-0.013$, $+0.497$] \\
\midrule
2L dense & (1,4) & 9 & $0.138\pm0.002$ & $0.152\pm0.002$ & $0.006\pm0.001$ & $0.010\pm0.001$ & $0.012\pm0.000$ & $0.001\pm0.002$ & $+0.132$ [$+0.118$, $+0.146$] & $+0.145$ [$+0.141$, $+0.150$] \\
 & (2,4) & 10 & $0.187\pm0.008$ & $0.181\pm0.002$ & $0.011\pm0.002$ & $0.015\pm0.002$ & $0.017\pm0.001$ & $0.001\pm0.002$ & $+0.176$ [$+0.135$, $+0.217$] & $+0.170$ [$+0.157$, $+0.183$] \\
 & (1,4)+3f & 12 & $0.119\pm0.003$ & $0.125\pm0.002$ & $0.008\pm0.001$ & $0.010\pm0.001$ & $0.013\pm0.000$ & $0.001\pm0.002$ & $+0.111$ [$+0.095$, $+0.126$] & $+0.117$ [$+0.107$, $+0.127$] \\
 & (2,4)+2f & 12 & $0.177\pm0.007$ & $0.171\pm0.002$ & $0.012\pm0.001$ & $0.015\pm0.002$ & $0.017\pm0.001$ & $0.000\pm0.002$ & $+0.164$ [$+0.134$, $+0.195$] & $+0.158$ [$+0.145$, $+0.171$] \\
 & (4,4) & 12 & $0.228\pm0.010$ & $0.206\pm0.006$ & $0.019\pm0.003$ & $0.024\pm0.002$ & $0.026\pm0.002$ & $-0.000\pm0.002$ & $+0.209$ [$+0.154$, $+0.264$] & $+0.186$ [$+0.150$, $+0.222$] \\
 & (6,4) & 14 & $0.253\pm0.006$ & $0.223\pm0.004$ & $0.027\pm0.003$ & $0.031\pm0.002$ & $0.033\pm0.003$ & $-0.000\pm0.002$ & $+0.226$ [$+0.189$, $+0.264$] & $+0.196$ [$+0.169$, $+0.223$] \\
 & (8,4) & 16 & $0.247\pm0.007$ & $0.227\pm0.003$ & $0.033\pm0.002$ & $0.036\pm0.003$ & $0.037\pm0.004$ & $-0.001\pm0.002$ & $+0.214$ [$+0.176$, $+0.252$] & $+0.193$ [$+0.171$, $+0.215$] \\
 & (4,7) & 18 & $0.206\pm0.014$ & $0.175\pm0.007$ & $0.013\pm0.003$ & $0.017\pm0.001$ & $0.022\pm0.002$ & $-0.000\pm0.002$ & $+0.193$ [$+0.120$, $+0.265$] & $+0.162$ [$+0.120$, $+0.203$] \\
\midrule
1L dense & (1,4) & 9 & $0.156\pm0.005$ & -- & $0.010\pm0.005$ & -- & -- & $0.001\pm0.002$ & $+0.145$ [$+0.125$, $+0.165$] & -- \\
 & (2,4) & 10 & $0.220\pm0.004$ & -- & $0.016\pm0.005$ & -- & -- & $0.001\pm0.002$ & $+0.204$ [$+0.187$, $+0.220$] & -- \\
 & (1,4)+3f & 12 & $0.119\pm0.003$ & -- & $0.010\pm0.005$ & -- & -- & $0.001\pm0.002$ & $+0.108$ [$+0.089$, $+0.127$] & -- \\
 & (2,4)+2f & 12 & $0.196\pm0.003$ & -- & $0.016\pm0.005$ & -- & -- & $0.001\pm0.002$ & $+0.180$ [$+0.163$, $+0.197$] & -- \\
 & (4,4) & 12 & $0.264\pm0.003$ & -- & $0.025\pm0.005$ & -- & -- & $0.001\pm0.003$ & $+0.239$ [$+0.227$, $+0.251$] & -- \\
 & (6,4) & 14 & $0.275\pm0.002$ & -- & $0.032\pm0.005$ & -- & -- & $0.001\pm0.003$ & $+0.243$ [$+0.230$, $+0.256$] & -- \\
 & (8,4) & 16 & $0.278\pm0.003$ & -- & $0.038\pm0.005$ & -- & -- & $0.001\pm0.003$ & $+0.240$ [$+0.226$, $+0.253$] & -- \\
 & (4,7) & 18 & $0.256\pm0.004$ & -- & $0.019\pm0.005$ & -- & -- & $0.001\pm0.003$ & $+0.238$ [$+0.216$, $+0.259$] & -- \\
\bottomrule
\end{tabular}}
\end{table}

\paragraph{Theorem-matched capacity and positive feature maps.} Table~\ref{tab:cap} compares the per-item criterion of Figure~\ref{fig:capacity} with Theorem~\ref{thm:cap}'s all-items, unrescaled criterion ($\delta=0.25$, $\epsilon=0.05$). Extending to $d_k=1024$ gives $n^\ast/(\delta d_k)=0.38, 0.52, 0.69$ at $d_k=64, 256, 1024$ (local exponent $1.21$ at $256$--$1024$). Across $\delta\in\{0.1,0.25,0.5\}$ and $\epsilon\in\{0.01,0.05,0.2\}$, exponents span $0.92$--$1.32$ (fidelity) and $1.75$--$1.94$ (identification). Delta identification gives $1.48$--$1.58$, while decay saturates near $1/(1-\gamma)=50$. Under $\mathrm{normalize}(\mathrm{elu}(g)+1)$ on Gaussian inputs, mean overlap is $0.685$ at every $d_k$, collapsing Hebbian fidelity to $n^\ast=1$ ($d_k\le 256$) and identification to $2$--$5$; mean-centering restores spherical exponents within $0.08$.

\begin{table}[t]
\centering\scriptsize
\caption{Capacity exponents of one Hebbian state (sphere keys, $d_k=16$--$256$, $\delta=0.25$, $\epsilon=0.05$; 95\% memory bootstrap CIs).}
\label{tab:cap}
\begin{tabular}{@{}lccc@{}}
\toprule
Criterion & $n^\ast$ (256) & Exponent & Local (256--1024) \\
\midrule
Fidelity, 95\% items, rescaled (Fig.~\ref{fig:capacity}) & 63 & 1.22 [1.22, 1.26] & 1.11 \\
Fidelity, 95\% items, raw & 46 & 1.14 [1.14, 1.14] & 1.12 \\
Fidelity, all items, rescaled & 45 & 1.12 [1.12, 1.16] & 1.20 \\
Fidelity, all items, raw (Thm.~\ref{thm:cap}) & 33 & 1.08 [1.06, 1.08] & 1.21 \\
Identification, 95\% items & 2188 & 1.89 [1.88, 1.89] & -- \\
Identification, all items (Thm.~\ref{thm:cap}) & 996 & 1.83 [1.82, 1.86] & -- \\
\bottomrule
\end{tabular}
\end{table}

\paragraph{Override persistence.} Table~\ref{tab:override} (Section~\ref{sec:exp_ext}) reports analytic re-pairing fractions (Proposition~\ref{prop:repair}). In trained one-layer models ($5+5$ trials, probe $10+r\le 17$), persistence relative to $W$-only and no-override baselines is $0.689\pm0.001$ (Hebbian), $0.668\pm0.001$ (decay), $0.676\pm0.001$ (Oja), $0.676\pm0.002$ (softmax), $0.356\pm0.003$ (PH), and $0.065\pm0.003$ (delta), flat across $r\in\{0,2,4,6,7\}$. Paired against Hebbian: delta $-0.624$ $[-0.636, -0.611]$, PH $-0.333$ $[-0.348, -0.317]$, softmax $-0.013$ $[-0.027, +0.001]$.

\paragraph{Trained extinction and PH sensitivity.} Across in-range trial splits (probes at 15--19) in one-layer elementary-feature models, DeltaNet loses $3$--$33\%$, PH $8$--$18\%$, softmax $89$--$98\%$, and contiguity rules $\le 0.4\%$. Under $8+8$ dense test cues: 2L dense-trained delta $14.3\pm2.2\%$, PH $1.4\pm0.3\%$, softmax $4.7\pm0.2\%$; 2L elementary-feature delta $41.9\pm33.2\%$, PH $15.9\pm10.6\%$, softmax $95.6\pm1.0\%$; 1L dense-trained delta $7.9\pm0.3\%$, softmax $3.1\pm0.1\%$. Over the \PhSensN-point PH grid ($w\in\{2,4,8\}$, $b\in\{-2,-1,0\}$, $\eta\in\{0.3,0.7\}$, $\beta\in\{0.2,0.3\}$), analytic blocking spans $[\PhSensBlockLo,\PhSensBlockHi]$ (median \PhSensBlockMed) and extinction spans $[\PhSensExtLo,\PhSensExtHi]\%$.

\paragraph{Distribution-shift recall and PH diagnostics.} Table~\ref{tab:ph} (Section~\ref{sec:exp_ph}) reports full accuracy on regime-switch and mixed-reliability tasks; Tables~\ref{tab:p6} and \ref{tab:p6b} detail the oracle/slope controls at noise $0.35$ and the lower-noise ($0.10$) slow-trace validation.

\begin{table}[t]
\centering\scriptsize
\caption{Mixed reliability at noise $0.35$ (8 pairs, 10 paired seeds).}
\label{tab:p6}
\begin{tabular}{@{}lcc@{}}
\toprule
Write rule & Accuracy & $\Delta$ vs.\ DeltaNet [95\% CI] \\
\midrule
DeltaNet & $0.7602\pm0.0022$ & -- \\
PH, learned $w,b$ & $0.7585\pm0.0029$ & $-0.0017$ [$-0.0054$, $+0.0020$] \\
PH, $w>0$ learned & $0.7604\pm0.0024$ & $+0.0002$ [$-0.0014$, $+0.0017$] \\
PH, $(w,b)=(0.75,-1.5)$ & $0.7298\pm0.0038$ & $-0.0305$ [$-0.0396$, $-0.0214$] \\
PH, $(w,b)=(4,-1)$ & $0.7441\pm0.0010$ & $-0.0161$ [$-0.0217$, $-0.0106$] \\
Oracle rate & $0.8321\pm0.0034$ & $+0.0718$ [$+0.0682$, $+0.0755$] \\
\bottomrule
\end{tabular}
\end{table}

\begin{table}[t]
\centering\scriptsize
\caption{Mixed reliability at stable-cue noise $0.10$ (8 pairs, 10 paired seeds; $F$: share of oracle gain).}
\label{tab:p6b}
\begin{tabular}{@{}lccc@{}}
\toprule
Write rule & Accuracy & $\Delta$ vs.\ DeltaNet [95\% CI] & $F$ \\
\midrule
DeltaNet & $0.9418\pm0.0027$ & -- & 0 \\
PH, slow trace ($\eta_0=0.047$) & $0.9451\pm0.0030$ & $+0.0033$ [$+0.0025$, $+0.0042$] & 0.143 \\
PH, default trace & $0.9429\pm0.0026$ & $+0.0011$ [$+0.0006$, $+0.0017$] & 0.049 \\
Oracle rate & $0.9650\pm0.0031$ & $+0.0232$ [$+0.0219$, $+0.0245$] & 1 \\
\bottomrule
\end{tabular}
\end{table}

\paragraph{Multi-head partition and specialization diagnostics.} Over the $(H,d_{\mathrm{head}})$ grid, $\log$ state size explains $R^2=\GridRtwoStateDelta$ (DeltaNet) and \GridRtwoStateHebb{} (Hebbian) of $\log$ error variance, vs.\ \GridRtwoHeadsDelta/\GridRtwoHeadsHebb{} for $\log H$. In 4-head DeltaNet models evaluated at load 24 ($8{,}192$ items), ablating any active head increases error by $+0.063\pm0.003$ (state 1,024) and $+0.052\pm0.015$ (state 4,096), per-item contributions are uncorrelated ($-0.030\pm0.005$ and $+0.026\pm0.019$), and head outputs have low cosine overlap ($0.09\pm0.03$ and $0.22\pm0.09$).

\begin{figure}[t]
\centering
\includegraphics[width=0.90\textwidth]{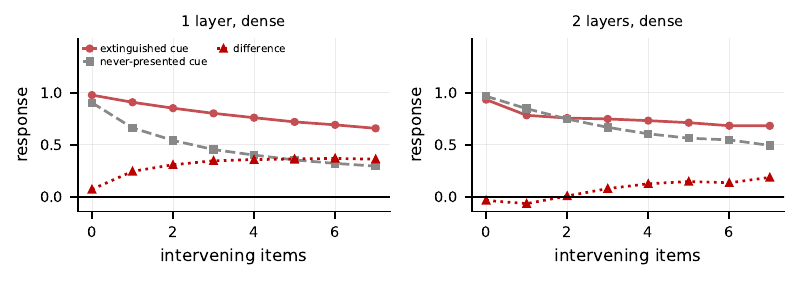}
\caption{Retention curves for DeltaNet (seed 0): the extinguished response decreases monotonically, while the never-presented-cue baseline decays faster, causing their difference to rise without actual recovery.}
\label{fig:recovery}
\end{figure}

\paragraph{Recovery, neutral fillers, ABA renewal, and $k$-hop depth.} Table~\ref{tab:recovery} and Figure~\ref{fig:recovery} show controlled retention curves across 60 models and 20 settings. In equivalence tests (5\% margin, 21 settings), 20 pass (5 of 6 with $\ge 10\%$ extinction); the sole exception is 2L elementary-feature DeltaNet (upper bound $+19.5\%$, driven by one seed rising $5.8\%$ toward baseline). With blank or zero-outcome cues instead of random fillers, 35 of 42 tests pass equivalence (all 1L elementary-feature pass; 6 of 12 2L elementary-feature exceed 5\%, max $+6.3\pm2.0\%$). In ABA renewal ($[(A{+}X_1)-(N{+}X_1)]-[(A{+}X_2)-(N{+}X_2)]$), 18 of 21 error-correcting models show no renewal beyond summation (three strongly extinguishing seeds show $+85.8$, $+62.4$, $+15$), whereas softmax renews $+38.8\pm0.9\%$ (1L) and $+43.8\pm0.9\%$ (2L) in elementary-feature models ($+14.6$ to $+32.2$ unnormalized dense). Table~\ref{tab:khop} reports $k$-hop depth accuracy, matching $L\ge k+1$ for $k\le 2$.

\begin{table}[t]
\centering\footnotesize
\caption{Spontaneous recovery in trained models with in-range probes and never-presented-cue controls (3 models $\times$ 5 protocol seeds). ``ext'': largest consecutive increase as \% of acquired response; ``spec'': paired difference $\text{ext}-\text{nov}$.}
\label{tab:recovery}
\setlength{\tabcolsep}{2.8pt}
\begin{tabular}{@{}lcccccccc@{}}
\toprule
& \multicolumn{2}{c}{Hebb} & \multicolumn{2}{c}{Decay} & \multicolumn{2}{c}{DeltaNet} & \multicolumn{2}{c}{Softmax} \\
\cmidrule(lr){2-3}\cmidrule(lr){4-5}\cmidrule(lr){6-7}\cmidrule(lr){8-9}
& ext & spec & ext & spec & ext & spec & ext & spec \\
\midrule
1L dense & $\RecLonedenseHebb$ ($\pm\RecSdLonedenseHebb$) & $\SpecLonedenseHebb$ & \multicolumn{2}{c}{--} & $\RecLonedenseDelta$ ($\pm\RecSdLonedenseDelta$) & $\SpecLonedenseDelta$ & $\RecLonedenseSoftmax$ ($\pm\RecSdLonedenseSoftmax$) & $\SpecLonedenseSoftmax$ \\
2L dense & $\RecLtwodenseHebb$ ($\pm\RecSdLtwodenseHebb$) & $\SpecLtwodenseHebb$ & $\RecLtwodenseDecay$ ($\pm\RecSdLtwodenseDecay$) & $\SpecLtwodenseDecay$ & $\RecLtwodenseDelta$ ($\pm\RecSdLtwodenseDelta$) & $\SpecLtwodenseDelta$ & $\RecLtwodenseSoftmax$ ($\pm\RecSdLtwodenseSoftmax$) & $\SpecLtwodenseSoftmax$ \\
1L elem. & $\RecLonesparseHebb$ ($\pm\RecSdLonesparseHebb$) & $\SpecLonesparseHebb$ & $\RecLonesparseDecay$ ($\pm\RecSdLonesparseDecay$) & $\SpecLonesparseDecay$ & $\RecLonesparseDelta$ ($\pm\RecSdLonesparseDelta$) & $\SpecLonesparseDelta$ & $\RecLonesparseSoftmax$ ($\pm\RecSdLonesparseSoftmax$) & $\SpecLonesparseSoftmax$ \\
2L elem. & $\RecLtwosparseHebb$ ($\pm\RecSdLtwosparseHebb$) & $\SpecLtwosparseHebb$ & $\RecLtwosparseDecay$ ($\pm\RecSdLtwosparseDecay$) & $\SpecLtwosparseDecay$ & $\RecLtwosparseDelta$ ($\pm\RecSdLtwosparseDelta$) & $\SpecLtwosparseDelta$ & $\RecLtwosparseSoftmax$ ($\pm\RecSdLtwosparseSoftmax$) & $\SpecLtwosparseSoftmax$ \\
\bottomrule
\end{tabular}
\end{table}

\begin{table}[t]
\centering\small
\caption{$k$-hop accuracy across depth $L$ (3 seeds). Both architectures match $L\ge k+1$ for $k\le 2$.}
\label{tab:khop}
\setlength{\tabcolsep}{5pt}
\begin{tabular}{@{}lcccccccc@{}}
\toprule
& \multicolumn{4}{c}{softmax} & \multicolumn{4}{c}{DeltaNet} \\
\cmidrule(lr){2-5}\cmidrule(lr){6-9}
$k$ & $L{=}1$ & $L{=}2$ & $L{=}3$ & $L{=}4$ & $L{=}1$ & $L{=}2$ & $L{=}3$ & $L{=}4$ \\
\midrule
1 & \KhopLoneHone & \KhopLtwoHone & \KhopLthreeHone & \KhopLfourHone & \KhopDLoneHone & \KhopDLtwoHone & \KhopDLthreeHone & \KhopDLfourHone \\
2 & \KhopLoneHtwo & \KhopLtwoHtwo & \KhopLthreeHtwo & \KhopLfourHtwo & \KhopDLoneHtwo & \KhopDLtwoHtwo & \KhopDLthreeHtwo & \KhopDLfourHtwo \\
3 & \KhopLoneHthree & \KhopLtwoHthree & \KhopLthreeHthree & \KhopLfourHthree & \KhopDLoneHthree & \KhopDLtwoHthree & \KhopDLthreeHthree & \KhopDLfourHthree \\
\bottomrule
\end{tabular}
\end{table}

\paragraph{Overshadowing and savings.} Measured relative to an element-alone control ($B$ after $AB{+}$ vs.\ $B{+}$ alone; salience ratio $3$, 5 seeds $\times$ 512), error-correcting recurrences overshadow strongly (delta $0.135\pm0.001$, PH $0.273\pm0.006$), whereas contiguity rules ($0.990\pm0.004$) and softmax ($1.000$) do not. Within-compound salience ratios reflect key scale directly (\OverHebb$\times$ Hebbian, \OverDecay$\times$ decay, \OverOja$\times$ Oja, \OverDelta$\times$ delta, \OverPh$\times$ PH, \OverSoftmax$\times$ softmax). On one-trial re-acquisition after extinction, DeltaNet recovers \SavingsDelta$\times$ a fresh cue (no savings, confirming erasure), versus \SavingsHebb$\times$ (Hebbian) and \SavingsDecay$\times$ (decay).

\FloatBarrier
\bibliographystyle{plainnat}
\bibliography{references}

\end{document}